%% file: main.tex
\documentclass[pmlr,twocolumn]{jmlr}

\usepackage{longtable}
\usepackage{csvsimple}
\usepackage{multirow}

 \usepackage{booktabs}

\usepackage[load-configurations=version-1]{siunitx} 

\usepackage{hyperref}
\usepackage{graphicx}
\usepackage{algorithm}

\newcommand{\joelcomment}[1]{{}}
\newcommand{\briancomment}[1]{{}}
\newcommand{\erancomment}[1]{{}}

\newcommand{\optumname}{Optum Chart }
\newcommand{\optumshortname}{Optum }

\makeatletter
\def\set@curr@file#1{\def\@curr@file{#1}} 
\makeatother

\theorembodyfont{\upshape}
\theoremheaderfont{\scshape}
\theorempostheader{:}
\theoremsep{\newline}

\jmlrvolume{193}
\firstpageno{1}

\jmlryear{2022}
\jmlrworkshop{Machine Learning for Health (ML4H) 2022}

\title[Extend and Explain]{Extend and Explain: Interpreting Very Long Language Models}

\author{\Name{Joel Stremmel}\Email{joel$\_$stremmel@optum.com}\\
  \Name{Brian L. Hill}\Email{brian.l.hill@optum.com}\\
  \Name{Jeffrey Hertzberg}\Email{jeffrey.hertzberg@optum.com}\\
  \Name{Jaime Murillo}\Email{jaime$\_$murillo@uhg.com}\\
  \Name{Llewelyn Allotey}\Email{llewelyn.allotey@optum.com}\\
  \Name{Eran Halperin}\Email{eran.halperin@optum.com}\\
  \addr Optum Labs, Minnetonka, MN, USA}

\begin{document}

\maketitle

\begin{abstract}

 
 While Transformer language models (LMs) are state-of-the-art for information extraction,
 long text introduces computational challenges requiring suboptimal preprocessing steps or alternative model architectures.
 Sparse attention LMs can represent longer sequences, overcoming performance hurdles.
 However, it remains unclear how to explain predictions from these models, 
 as not all tokens attend to each other in the self-attention layers, and 
 long sequences pose computational challenges for explainability algorithms 
 when runtime depends on document length.
 These challenges are severe in the medical context where documents can be very long, and 
 machine learning (ML) models must be auditable and trustworthy.
 We introduce a novel Masked Sampling Procedure (MSP) to identify the text blocks that contribute
 to a prediction, apply MSP in the context of predicting diagnoses from medical text, and validate our approach with a blind review by two clinicians.
 Our method identifies $\approx 1.7\times$ more clinically informative text blocks than the previous state-of-the-art,
 runs up to $100\times$ faster, and is tractable for generating important phrase pairs.  
 MSP is particularly well-suited to long LMs but can be applied to any text classifier.
 We provide a general implementation here.\footnote{ \url{https://github.com/Optum/long-medical-document-lms} }


\end{abstract}


\section{Introduction}

During a visit to a medical care provider, a physician typically records important information about patient presentation,
diagnosis, and treatment via free-form text.
These notes contain rich clinical data not found in structured electronic medical records.
For example, information about patient experience and even some diseases may not be documented with standardized codes,
but evidence of symptoms or diagnoses may be recorded in free-form text.
Clinical notes can also prove useful for extracting relevant medical conditions for cohort building,
such as for identifying patients for clinical trials or developing disease progression models using features from the text.
Because these text sequences can be very long, digging through patient records to find relevant information is a
tedious, manual process, and automation with standard machine learning (ML) approaches is often insufficient.
This motivates the development of an accurate, automated, and interpretable method for extracting medical conditions from long medical documents.



Convolutional neural network (CNN) models have been the architecture of choice for
long medical text \citep{mullenbach_explainable_2018,gehrmann_comparing_2018,li_icd_2019,reys_predicting_2020,hu_explainable_2021},
largely due to the computational complexity of the self-attention mechanism in Transformers like BERT \citep{devlin_bert_2019}. 
However, recent advancements in sparse attention or long language models (LMs) \citep{zaheer_big_2021,choromanski_rethinking_2021,beltagy_longformer_2020,kitaev_reformer_2020} 
suggest it is now possible to represent long medical documents without convolutions that fail to capture interactions between distant tokens in a text sequence, 
or the truncation and segmentation with pooling methods that ML practitioners apply to standard Transformers in practice \citep{huang_clinicalbert_2020}.

While there are approaches for interpreting the predictions of traditional ML models
and neural networks \citep{sundararajan_axiomatic_2017,lundberg_unified_2017},
understanding the blocks of text driving the predictions of long LMs is not straightforward.
A common approach to interpret the predictions of Transformers is to examine the attention weights of tokens.
Although subword tokenization has been shown to be performant in downstream classification tasks, the attention weights of individual subword tokens
are not always informative or interpretable, and attention weight interpretation has been criticized \citep{jain_attention_2019}.
The limitations of word and subword level explanations are especially prevalent in a healthcare context where word pieces are often
divorced of clinical meaning or capture only part of a phrase representing a clinical concept.
For example, consider the phrase, ``patient developed atrial fibrillation," consisting of many tokens.
Understanding the impact of blocks of text (represented as sequences of subword tokens) on model predictions 
only becomes more challenging for models with sparse attention, as not all subwords attend to each other in the self-attention layers.

In this work, we introduce a novel method, the Masked Sampling Procedure (MSP), 
to identify important blocks of text used by long LMs or any text classifier to predict document labels.
Our method is unique and valuable in that we simultaneously mask multi-token blocks to answer the counterfactual: “what if this block of text had been absent?”
Unlike previous work,
runtime does not depend on document length, 
and we provide a rigorous method to compute p-values for each text block.
Our method extends to any number of multi-token text blocks to measure interactions, 
and we report the benefits 
of specific masking probabilities.

We validated that MSP returns clinically informative explanations of 
medical condition predictions from a very long LM with a blinded experiment involving two physicians. 
In Section~\ref{sec:msp_validation} we share insights from our clinician collaborators 
regarding the explanations surfaced by MSP and describe the 
superior performance and runtime efficiency of MSP compared to the state-of-the-art,
showing that our method is up to $100\times$ faster and $\approx 1.7\times$ better at identifying important text blocks 
from a very long LM applied to long medical documents.
Finally, in Section~\ref{sec:classifier_performance}, 
we describe the benefit of using sparse attention LMs in the context of predicting medical conditions from
long medical documents (up to 32,768 tokens),
extending the length of the typical LM for clinical documents from 512 to 32,768 tokens
with an over 5\% absolute improvement in
micro-average-precision over a popular and effective CNN architecture for 
predicting medical conditions from text on four different size train sets.

\section{Related Work}


Many methods have been proposed to explain the predictions of text classifiers, such as those that examine
the individual attention weights of tokens in LMs \citep{skrlj_exploring_2021}, 
gradient-based methods that attempt to reveal the saliency of individual tokens \citep{yin_interpreting_2022},
and approaches that perturb the input text to measure importance \citep{kokalj_bert_2021}.
The most similar approach to our procedure is likely the Sampling and Occlusion (SOC) algorithm \citep{jin_towards_2020}.
\citet{jin_towards_2020} apply SOC to BERT and show that SOC outperforms a variety
of competitive baselines including GradSHAP \citep{lundberg_unified_2017}, a popular approach
combining ideas from Integrated Gradients \citep{sundararajan_axiomatic_2017} with perturbation-based
feature importance, on three benchmark datasets.  
SOC masks one word or text block at a time to compute the impact on label predictions and eliminates 
the dependence on surrounding context for a given block by sampling neighboring words from a trained LM.
However, if the trained LM performs well, the sampled neighboring words will be similar to the original context. 
This sampling procedure is computationally expensive which we discuss in Section~\ref{sec:runtime_comparison}.

Further related work includes traditional text representation approaches
like TF-IDF~\citep{sparck_jones_statistical_1988} and word2vec~\citep{mikolov_efficient_2013},
predicting medical conditions using CNNs with attention~\citep{mullenbach_explainable_2018, hu_explainable_2021, lovelace_dynamically_2020}, 
LMs that improve on traditional text representations~\citep{vaswani_attention_2017, devlin_bert_2019},
sparse attention LMs~\citep{kitaev_reformer_2020, beltagy_longformer_2020, choromanski_rethinking_2021, zaheer_big_2021}, 
and domain-specific pretraining of LMs for medical text~\citep{alsentzer_publicly_2019, lee_biobert_2020, liu_self-alignment_2021}.
See Supplemental Section~\ref{sec:further_discussion_of_related_work} for more details.

To our knowledge, the only research applying long LMs to the clinical domain is ~\citet{li_clinical-longformer_2022}.
The authors fine-tuned Longformer and Big Bird~\citep{zaheer_big_2021} for clinical question answering~\citep{pampari_emrqa_2018} and named-entity recognition.
We benchmark the first clinically pretrained long LM for multi-label classification of conditions from clinical notes, extending the typical sequence length of
long LMs by $8\times$ (from 4,096 to 32,768 tokens) to address the problem of extracting information from long, individual, patient medical histories.

\section{Cohort}


We use two document types in our experiments: medical charts, which are long-form clinical notes concatenated from many visits, and discharge summaries.
In both cases, we consider a single document to be the entire sequence of tokens.

The \optumname dataset consists of 6,526,116 full-length medical charts for Medicare patients from 2017-2018
and was obtained by Covered Entity customers of Optum Labs to provide quality improvement services.
We used 5,481,937 unlabeled charts for pretraining text representations.
We used 640,000 labeled charts for training, 64,000 for validation, and 187,953 for testing.
Labels were generated by human medical coders in a process wherein three coders had to agree on each medical condition label before assigning it. 
Data split sizes were determined to ensure a fair comparison to existing models
and generating these final splits involved downsampling to measure the effect of training set size
and reduce evaluation time on the validation set during training.
Specifically, we downsampled at random from 730,925 train samples and 125,301 validation samples 
while maintaining an initial split of 187,953 testing samples.  These original 
split sizes were the result of a multi-label, stratified shuffle split using iterative stratified sampling \citep{sechidis_stratification_2011}
with target proportions of 70\%, 12\%, and 18\% respectively. 
This splitting procedure was used to ensure a roughly uniform distribution of labels across train, validation, and test splits. 
The remaining, unlabeled charts not included in this splitting procedure were used for pretraining.
While we broke the train set into four datasets to measure the effect of train set size, 
the same validation and test sets were used in all experiments.
The median length in subword tokens of documents in the \optumname dataset is 4,043 tokens with interquartile range [1,830 - 9,142].
Descriptive statistics can be found in Supplemental Table~\ref{tab:descriptive_stats} and 
condition prevalence in Supplemental Table~\ref{tab:optumcharts85_performance_compare}.

MIMIC-III \citep{johnson_mimic-iii_2016} contains de-identified clinical records for
intensive care unit (ICU) patients treated at Beth Israel Deaconess Medical Center.
Included is a set of discharge summary notes and International Classification of Disease ICD-9 diagnoses associated with each ICU stay.
We use the subset of discharge summaries from \citet{mullenbach_explainable_2018} consisting of
11,371 notes from 2001-2012 and the top 50 most common ICDs appearing in each summary.
The median length in subword tokens of documents in this dataset is 1,430 tokens with interquartile range [1,029 - 1,929].
Descriptive statistics can be found in Supplemental Table~\ref{tab:descriptive_stats}.
We use the same 8,067 sample train, 1,574 validation, and 1,730 test sets as in \citet{mullenbach_explainable_2018}.

\section{Methods}





\subsection{Masked Sampling Procedure}

\begin{figure*}[h]
  \centering
  \includegraphics[width=6.0in]{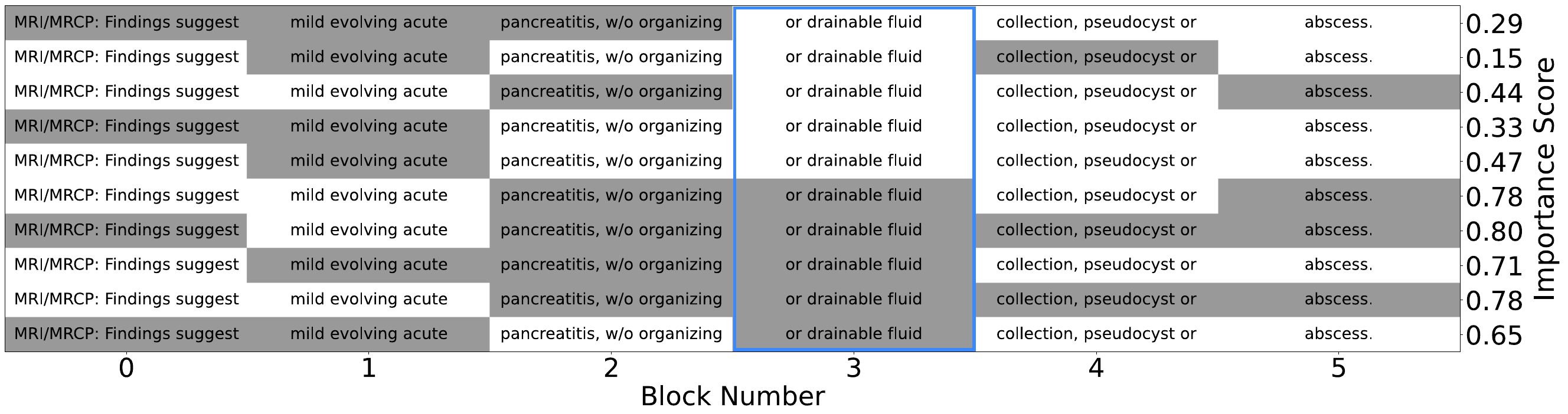}
  \caption{Diagram describing MSP input and output. 
  Each row is an example medical text sequence, where some number of text blocks are randomly masked (shown in gray) and used as input to a classifier. 
  For each masked row we get an importance score, defined as the difference between the baseline prediction, when no text blocks are masked, and the prediction when blocks are masked.
  For a specific text block of interest (e.g., block 3 in blue), we then calculate the mean difference in importance scores when the block is/is not masked to measure the contribution of that text block to the prediction.
  }
  \label{fig:msp_diagram}
\end{figure*}

To reveal which text blocks have the largest effect on the predictions of long LMs or any text classifier,
we propose MSP (Algorithm~\ref{alg:masked_sampling}).
To explain predictions from a text sequence, MSP randomly masks all text blocks of size $B$ subwords with probability $P$,
feeds the new sequence to the classifier, then measures the difference in label probability between the masked and unmasked versions of the sequence (see Figure~\ref{fig:msp_diagram}).
Over many iterations $N$, large differences in predicted probabilities originating from masking a given text block
suggest the block contributed important evidence to the label prediction.
MSP outputs the top $K$ most important blocks for each label along with a measure of statistical significance
computed by comparing to randomly sampled text blocks using a bootstrap procedure, with the null hypothesis, that,
text blocks with high importance, as determined by MSP, are no more important to a label prediction
than randomly sampled blocks (see Algorithm~\ref{alg:masking_sig}).

\SetKwComment{Comment}{/* }{ */}

\begin{algorithm2e}[ht]
\caption{Masked Sampling Procedure (MSP)}\label{alg:masked_sampling}
\DontPrintSemicolon
\KwData{$X_{i} \in \mathbb{R}^{S_{i} \times d_{c}}$} 
\KwResult{maskedSampleProbs $\in \mathbb{R}^{N \times L}$, maskIndices $\in \{0,1\}^{N \times (S_{i}/B)}$}
\SetKwFunction{FSum}{MSP}
\SetKwProg{Fn}{Function}{:}{}
\SetKw{KwBy}{by}
\Fn{\FSum{$X$, $N$, $B$, $P$}}{
  maskedSampleProbs $\gets$ [ ]\;
  maskedIndices[$1:N, 1:(S_{i}/B)$] $\gets 0$\;
  $\hat{y_{i}} \gets$ Classifier($X_{i}$)\;
  \For{$n=1$ \KwTo $N$}{
    \For{$j=1$ \KwTo $S_{i}$ \KwBy B}{
      $r \gets \mathcal{U}(0, 1)$\;
      \If{$r < P$}{
        $X[j:j+B] \gets$ maskToken\;
        maskedIndices[n, j/B] = 1\;
      }
    }
    $\hat{y_{n}} \gets$ Classifier($X_{i}$)\;
    $\Delta \hat{y_{n}} \gets \hat{y_{i}} - \hat{y_{n}}$\; 
    maskedSampleProbs.append($\Delta \hat{y_{n}}$)
  }
  \KwRet maskedSampleProbs, maskedIndices\;
}
\end{algorithm2e}

In a blind experiment, two clinicians validated the ability of MSP to explain predictions from a 
very long Big Bird model (Section~\ref{sec:very_long_big_bird}) on randomly sampled discharge summaries from MIMIC
compared to SOC~\citep{jin_towards_2020} and a random algorithm.  
We selected the very long Big Bird model for its ability to represent long medical documents
and because attention weight analysis of this model is not straightforward due to the sparse self-attention mechanism.
Each clinician received the same 400 text block-diagnosis pairs from each algorithm 
and independently annotated the text blocks as either uninformative or informative for making
the ICD diagnosis.
Each of the 1,200 total text blocks supplied to each clinician were among the top five most important
for the corresponding label according to MSP, SOC, or by random selection
(see Supplemental Methods~\ref{sec:blind_experiment_sampling} for more details on how text blocks were selected).
We compared the number of informative text blocks from each method along with differences in runtime.

For MSP, we set $P=0.1$ according to an experiment with a single clinical reviewer comparing values of $P$ shown in
Supplemental Table~\ref{tab:masking_experiment}.  For a fair comparison to SOC, we fixed $B=10$ and set the expected number of times a given phrase is masked
to 100.  We used the sampling radius of 10 tokens recommended by \citet{jin_towards_2020} and set the number of sampling rounds to
100.


\subsection{Baseline Text Representations}

We compared the performance of several text representations and classifiers for the task of predicting medical conditions from clinical text 
to the Big Bird LM for which we generated explanations with MSP. These methods operated at either the word or subword level
following text preprocessing (see Supplemental Methods~\ref{sec:dataset_preprocessing}).
More details on baseline text representations can be found in Supplemental Methods~\ref{sec:appendix_model_development_language}

\subsection{Very Long Big Bird}
\label{sec:very_long_big_bird}


Big Bird's sparse attention mechanism approximates full self-attention with a combination of
global tokens, sliding window attention, and random approximations of fully connected graphs representing full self-attention.
These mechanisms take the memory consumption from $O(L^2)$ to $O(kL)$, where $k$ is the size of the sliding attention window.
To pretrain a Big Bird LM on clinical text, we first trained a Byte Pair Encoding subword tokenizer \citep{sennrich_neural_2016}
to tokenize the text.  This same tokenization approach was used by ~\citet{zaheer_big_2021}.
After cleaning, we truncated all text to 32,768 subwords following tokenization,
and pretrained with masked language modeling (MLM) as in~\citet{zaheer_big_2021}.
We selected 32,768 subword tokens to increase the maximum sequence length represented by Big Bird 
by another $8\times$, given that the original Big Bird model is $8\times$ the maximum sequence length of BERT (512 to 4,096).  
Furthermore, over 95\% of the medical charts in the \optumname dataset
are less than 32,768 tokens in length.

\subsection{Text Classifiers}

We are interested in identifying conditions from medical documents relevant to diagnosing or treating patients
and focused our experiments on two datasets with 85 and 50 medical condition labels respectively.
To predict these conditions, we used ElasticNet, a Feed Forward Neural Network (FFNN),
BERT variants with text segmentation and pooling (on MIMIC only),
CAML, and Big Bird, all trained as multi-label classifiers.
Here we describe CAML and Big Bird, which were the most competitive.
Details on all models can be found in Supplemental Methods~\ref{sec:appendix_model_development_classification}.

CAML uses a CNN
layer to extract features from the word2vec embedding matrix and an attention mechanism to localize signal for a particular prediction task.
We implemented CAML as described in \cite{mullenbach_explainable_2018} using a CNN layer with
filter size between 32 and 512, kernel size between 3 and 10, and dropout on the embedding layer between 0 and 0.5.
The output is a vector of probabilities, one for each label, to which we applied the
sigmoid function and trained the model to minimize binary cross-entropy loss.

The \optumname sequences are $8\times$ larger than the typical "long" sequence \citep{tay_long_2020} at a maximum length of 32,768 tokens.
We pretrained Big Bird from random initialization on medical documents, added a classification head with
a single feed-forward layer of size 1,536 (2x the hidden size), an output layer with one neuron per label, and trained using binary cross-entropy loss.

\section{Results}




\subsection{Clinical Validation of MSP}
\label{sec:msp_validation}

We examine the clinical utility of MSP in a blind experiment with two clinicians, first
discussing examples of informative text blocks, then comparing the number of
informative text blocks surfaced by MSP to the SOC algorithm in the blind experiment.
Finally, we discuss runtime.

\subsubsection{Informative Text Blocks}


\input{tables/masked_text_results.tex}

Table~\ref{tab:masked_text_results} depicts example text blocks and their importance computed via MSP that were
deemed informative during an initial clinical review.
This review confirmed three general features that drive text block “informative-ness.”

The most obvious were exact matches with diagnosis text.
For example the text block “pneumonia patient being discharged o n maximal copd regimen including,”
was highly informative in implying a diagnosis of “pneumonia, organism unspecified.”
Less obvious were synonyms or close synonyms for a diagnosis.
In the text-string “lovenox bridge nstemi o n admission the patient had elevated,”
``NSTEMI" is an acronym for ``non-ST segment elevation myocardial infarction," which is synonymous with the diagnosis ``subendocardial infarction."

Other common elements in highly informative blocks were drugs that are always, or almost always used for a particular diagnosis.
MSP identified the block “consulted amiodarone was held rhythm slowly began to recover she,”
associated with the diagnosis “atrial fibrillation.”
Amiodarone is an antidysrhythmic drug mostly used for atrial fibrillation.

MSP also identified obscure but clinically relevant blocks, such as “al likely improve as pna improves s p cabg complicated,”
including “s”, “p”, and “cabg.” Grouped together, these suggest the patient is “status-post” coronary artery bypass grafting,
meaning they have had the procedure.
In order to graft coronary arteries, the patient must be placed on an aortocoronary bypass machine to allow the procedure to be completed.
This block was associated with the diagnosis of “aortocoronary bypass status.”
Another seemingly obscure but clinically informative block, ``albuterol and ipra prn his acidosis slowly improved as did"
appeared for the diagnosis of ``acute respiratory failure."
Even though none of the words comprising the diagnosis exist in the block, clinician review confirmed that the block is associated
with acute respiratory failure, despite the lack of matches for words in the diagnosis label.
Albuterol, a fully and correctly spelled-out drug associated with respiratory distress,
is related to bronchial obstruction, seen in chronic obstructive pulmonary disease (COPD).
Ipra, an abbreviation for ipratropium bromide, is used in COPD. COPD is a common cause of acute respiratory failure.
Acidosis, identified through arterial blood testing, is a sign of hypoventilation, which causes elevation in blood carbon dioxide levels and resultant accumulation of H\textsubscript{2}CO\textsubscript{3} (an acid).
This is seen in people with COPD exacerbation who experience respiratory failure.

\subsubsection{Blind Experiment Analysis}

\input{tables/msp_vs_soc_vs_rand.tex}

\begin{figure}[h]
  \centering
  \includegraphics{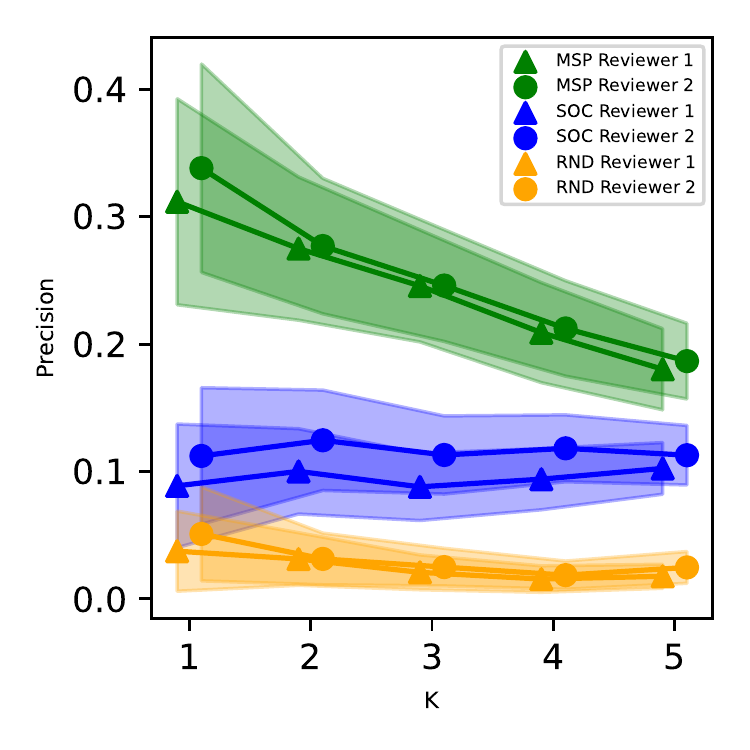}
  \caption{Precision for the top $K$ text blocks surfaced by MSP, SOC, and the random algorithm (RND) according to each reviewer
  for each document-label pair with 95\% confidence intervals computed using 1000 bootstrap iterations.}
  \label{fig:ps_at_k}
\end{figure}

Two clinicians received 400 text block-diagnosis pairs from each of MSP, SOC, and the random algorithm 
and independently annotated the text as either uninformative or informative for making
the ICD diagnosis.
Table~\ref{tab:msp_vs_soc_vs_rand} depicts the number of informative text blocks surfaced by each explainability algorithm.
Figure~\ref{fig:ps_at_k} depicts the precision of each algorithm according to both reviewers.  
MSP is superior to SOC in terms of the total number of clinically informative text blocks surfaced and precision, 
especially when limiting the number of blocks surfaced to a small number.
Supplemental Figure~\ref{fig:mrrs_at_k} depicts performance of these algorithms from an information retrieval perspective.




\subsubsection{Runtime Comparison}
\label{sec:runtime_comparison}

\input{tables/algo_runtimes.tex}
\input{tables/algo_runtimes_theoretical.tex}

On the MIMIC discharge summaries of modest length (IQR 1,029-1,929),
MSP was up to $100\times$ faster than SOC (Table~\ref{tab:algo_runtimes}).
For $J$ sampling iterations per block, masking probability $P$, and document length $L$,
using MSP, the number of evaluations of the text by the classifier is $O(J/P)$ for computing the importance of individual sentences and $O(J/P^2)$
for pairs.  Using SOC, the number of evaluations is $O(JL)$ and $O(JL^2)$ respectively.
Thus, the run time of our approach does not grow with the document length as the number of model evaluations does not depend on $L$.
Since SOC has a quadratic dependency on $L$, it is
very expensive for computing the importance of individual sentences in documents of even modest length and
infeasible for computing the importance of sentence pairs (see example in Table~\ref{tab:algo_runtimes_theoretical}).
In the medical and other domains, we expect distant pieces of information to interact,
and use pairs analysis with MSP to
demonstrate how Big Bird integrates distant contextual information in Supplemental Results~\ref{sec:integrating_distant_contextual_information}.



\subsection{Medical Condition Prediction}
\label{sec:classifier_performance}



\input{tables/model_perf_compare.tex}

We assessed model performance when predicting medical conditions in long medical charts from the \optumname dataset.
Since the prevalence of each label is often very low (median: 0.6\%), we used precision and recall as our metrics of interest (specifically, area-under the precision-recall (AUPR) curve, or average-precision (AP)) \citep{saito_precision-recall_2015}.
In Table~\ref{tab:model_perf_compare} we show performance in terms of AP micro- and macro-averaged across labels.
For most labels, Big Bird outperformed CAML, (see Supplemental Figure~\ref{fig:best_model_comparison}a), 
and across training datasets of four sizes performed over 5\% better than CAML in micro-average-precision (see Supplemental Results~\ref{sec:effect_of_training_set_size} and Supplemental Tables~\ref{tab:12800_perf_transpose}, \ref{tab:64000_perf_transpose}, \ref{tab:128000_perf_transpose}, \ref{tab:640000_perf_transpose}).
Supplemental Figure~\ref{fig:best_model_comparison}c shows
the $\textnormal{log}_{2}$-scaled ratio of the Big Bird AUPR to the CAML AUPR as a function of label prevalence.
Big Bird generally outperforms CAML on labels with prevalence $>$ 5\%,
but many of the most significant improvements are found in rare labels (prevalence $\leq$ 5\%).
AUPRs for each label are included in Supplemental Table~\ref{tab:optumcharts85_performance_compare}.


Next, we assessed performance for predicting any of the 50 most common ICD-9s assigned to a MIMIC discharge summary.
As baselines, we trained multiple TF-IDF-based models, CAML, and several BERT variants with different types of pooling over segments of text (see Supplemental Methods~\ref{sec:appendix_model_development_classification}).
We explored Big Bird architectures with varying sequence lengths (4,096 or 32,768 tokens) and pretraining datasets (general or clinical text).
As shown in Table~\ref{tab:model_perf_compare}, on MIMIC, with clinical pretraining, Big Bird with sequence length 4,096 slightly outperformed Big Bird with sequence length 32,768. 
This is likely due to the average document length in MIMIC being shorter than 4,096 tokens.
We found that Big Bird (4,096 max sequence length) pretrained on the \optumname dataset outperformed Big Bird (4,096 max sequence length) pretrained on generic, English text (Supplemental Table~\ref{tab:8067_perf_transpose}).
This supports previous work demonstrating that in-domain pretraining from scratch is superior to cross-domain fine-tuning for tasks in the biomedical domain~\citep{lee_biobert_2020}.

Of all the baselines we compared with various flavors of Big Bird, CAML performed best on MIMIC.
Supplemental Figure~\ref{fig:best_model_comparison}b and d shows the performance of Big Bird and CAML for each label. 
BERT variants using pooled segment representations performed worse than CAML and Big Bird with clinical pretraining (Supplemental Table~\ref{tab:bert_mimic_results}). 
This suggests learning a representation of an entire input sequence with sparse self-attention outperforms aggregation over segments.

\section{Discussion}




The purpose of this research is to extract meaningful insights from long medical documents 
in an auditable and transparent way.  
We discussed and demonstrated the performance benefits of 
sparse attention LMs for extracting medical conditions from very long text
and proposed MSP to address the major challenge of interpreting long LM predictions.  
MSP can explain medical condition predictions from discharge summaries using the very long Big Bird LM $\approx 1.7\times$ better than a state-of-the-art 
explainability algorithm and up to $100\times$ faster.  It's also tractable for generating important text block pairs.

We note that among other limitations of our research, medical charts contain more than just free-form text. 
Some semi-structured information from tables and bulleted lists is lost when representing an entire chart as a single text sequence. 
Discharge summaries, as in the MIMIC dataset, are just one type of medical note, specific to an inpatient setting.
The \optumname dataset is more comprehensive in that it consists of full length medical charts, 
but these charts are only for Medicare patients.
Future work should examine other types of clinical notes as well as other populations.

Additional opportunities for further research include
modifying the pretraining procedure for long LMs applied to medical text to take advantage of additional information available in electronic health records, 
evaluating a highlighting system that surfaces information relevant to diagnosing and treating patients, 
and assessing long LMs for clinical information extraction for bias using methods like MSP.

We view improving the underlying representations of medical text and understanding the predictive elements as key steps toward ensuring that
ML can be safely deployed in the medical domain but acknowledge there is more work to do
to scale ML across the healthcare system in a just and transparent way.

\clearpage
\bibliography{stremmel22}

\clearpage
\appendix

\section{Further Discussion of Related Work}
\label{sec:further_discussion_of_related_work}

We consider TF-IDF~\citep{sparck_jones_statistical_1988} and word2vec~\citep{mikolov_efficient_2013} as foundational
approaches for representing text and use these methods as baselines.
Both \citet{kim_convolutional_2014} and \citet{mullenbach_explainable_2018} use pretrained word2vec embeddings
to provide intelligent initialization to the embedding layer of one-dimensional CNN models that process text sequences.
\citet{mullenbach_explainable_2018} propose the CAML architecture which introduces
an attention mechanism in the CNN and focuses on the task of predicting medical conditions from text as
multi-label classification. \citet{hu_explainable_2021} propose a wide version of CAML, and we incorporate this
recommendation in our implementation, grid searching for the number of filters (up to 512) and filter size (up to 10).
\cite{lovelace_dynamically_2020} modify the CAML architecture to adapt it to the task of jointly predicting both medical codes
and subsequent patient outcomes like readmission and mortality.

LMs have yet to be widely adopted for the task of predicting medical conditions from text,
though architectures such as the Transformer~\citep{vaswani_attention_2017} and BERT~\citep{devlin_bert_2019} have revolutionized
the way ML practitioners classify documents, outperforming CNN-based approaches on generic, English
classification tasks~\citep{sun_how_2020}.
We observe two main challenges with applying Transformer architectures like BERT to medical documents:
first, medical documents differ from generic, English text,
and second, medical documents containing entire patient medical histories can be very long, far exceeding the maximum sequence length used by BERT.

The Clinical BERT~\citep{alsentzer_publicly_2019}, BioBERT~\citep{lee_biobert_2020}, and SapBERT~\citep{liu_self-alignment_2021} architectures
were all designed to tackle the important differences between generic, English natural language and text encountered in the clinical and biomedical domains.
Clinical BERT further pretrains the standard BERT model on medical notes from MIMIC-III, though MIMIC-III is limited in the total number of documents.
The BioBERT architecture solves this problem by training on more samples and demonstrates that in-domain pretraining from scratch,
outperforms continued pretraining on data from the BLURB biomedical benchmark.
That said, the domain of biomedical papers used to pretrained BioBERT differs from the domain of individual, patient medical histories
and the task of predicting specific conditions from text.
One solution is to incorporate more information about synonym and subtype relationships found in medical language.
The SapBERT authors demonstrate the value of this approach through a continued pretraining strategy whereby the BERT model learns to align
entities in the Unified Medical Language System. All of these approaches demonstrate benefits over base BERT,
but none are well-suited to long documents due to BERT's quadratic time and memory complexity.

To address the problem of long documents, sparse self-attention architectures have been proposed that approximate BERT's self-attention including
Reformer~\citep{kitaev_reformer_2020}, Longformer~\citep{beltagy_longformer_2020}, Performer~\citep{choromanski_rethinking_2021}, and Big Bird
~\citep{zaheer_big_2021}.
Reformer replaces global self-attention by locality-sensitive hashing (LSH) self-attention based on similarity of the query vectors
in the $\textnormal{softmax}(QK^T)V$ self-attention equation.
This takes the attention time complexity from $O(L^2)$ to $O(L \textnormal{ log}(L))$ where $L$ is the length of the input sequence.
Longformer also applies self-attention over windows of nearby tokens and introduces a subset of global tokens that attend to every other token.
The Big Bird LM combines the ideas of local and global attention with a random attention mechanism using a random graph approximation
of fully connected graphs representing full self-attention.
Intuitively, Big Bird attempts to create a path between any two tokens in the input sequence with the limitation that many layers might be
required to connect any two tokens.  Because Big Bird combines multiple self-attention approximation strategies, provides an intuitive
implementation of sparse self-attention, and achieves the best average performance across six long document tasks reported on the Long-Range Arena (LRA)
benchmark \citep{tay_long_2020}, we focus on Big Bird in our research.
Luna~\citep{ma_luna_2021} and S4~\citep{gu_efficiently_2022} report higher performance numbers on LRA in their papers,
though, at the time of writing, these numbers have not yet been added to the LRA website.\footnote{ \url{https://github.com/google-research/long-range-arena/blob/09c2916c3f33a07347dcc70c8839957d3c9d4062/README.md} }

\section{Supplemental Methods}
\label{sec:supplemental_methods}

\subsection{Cohort Descriptive Statistics}

\input{tables/descriptive_stats.tex}

The preprocessing steps used by \citet{mullenbach_explainable_2018} for MIMIC discharge summaries are provided here.\footnote{ \url{https://github.com/jamesmullenbach/caml-mimic/blob/44a47455070d3d5c6ee69fb5305e32caec104960/notebooks/dataproc_mimic_III.ipynb} }

\subsection{Exploratory Data Analysis}
\label{sec:eda}

Descriptive statistics for the medical documents used in this research can be found in Table~\ref{tab:descriptive_stats}.

\subsection{Dataset Preprocessing}
\label{sec:dataset_preprocessing}

To clean the text of each document prior to tokenization, we strip these characters from the start and end of all words: \verb-.!"#$&'()*+,/:;?@[\]^_`{|}~-.
We then remove these characters entirely: \verb-!"#$&'()*+,;?@[\]^_`{|}~”-.
These characters are not relevant in a healthcare context in the same way as retained characters such as, for example, ``/", which is important for blood pressure readings.
We then eliminate words longer than 39 characters, characters occurring three times in a row within a word, dates, phone numbers, URLs, states, names, cities, and emails.
We leave numbers as they are (no bucketing).
Numbers appearing in a healthcare context are relatively small, and it has been shown that Transformer LMs can effectively
represent the relative size of numbers in the range of -500 to 500 \citep{wallace_nlp_2019}.

\subsection{Pretraining and Fine-Tuning Approach for Clinical Text}
\label{sec:pretraing_and_finetuning_approach}

\begin{figure*}[h]
  \centering
  \includegraphics[width=5.0in]{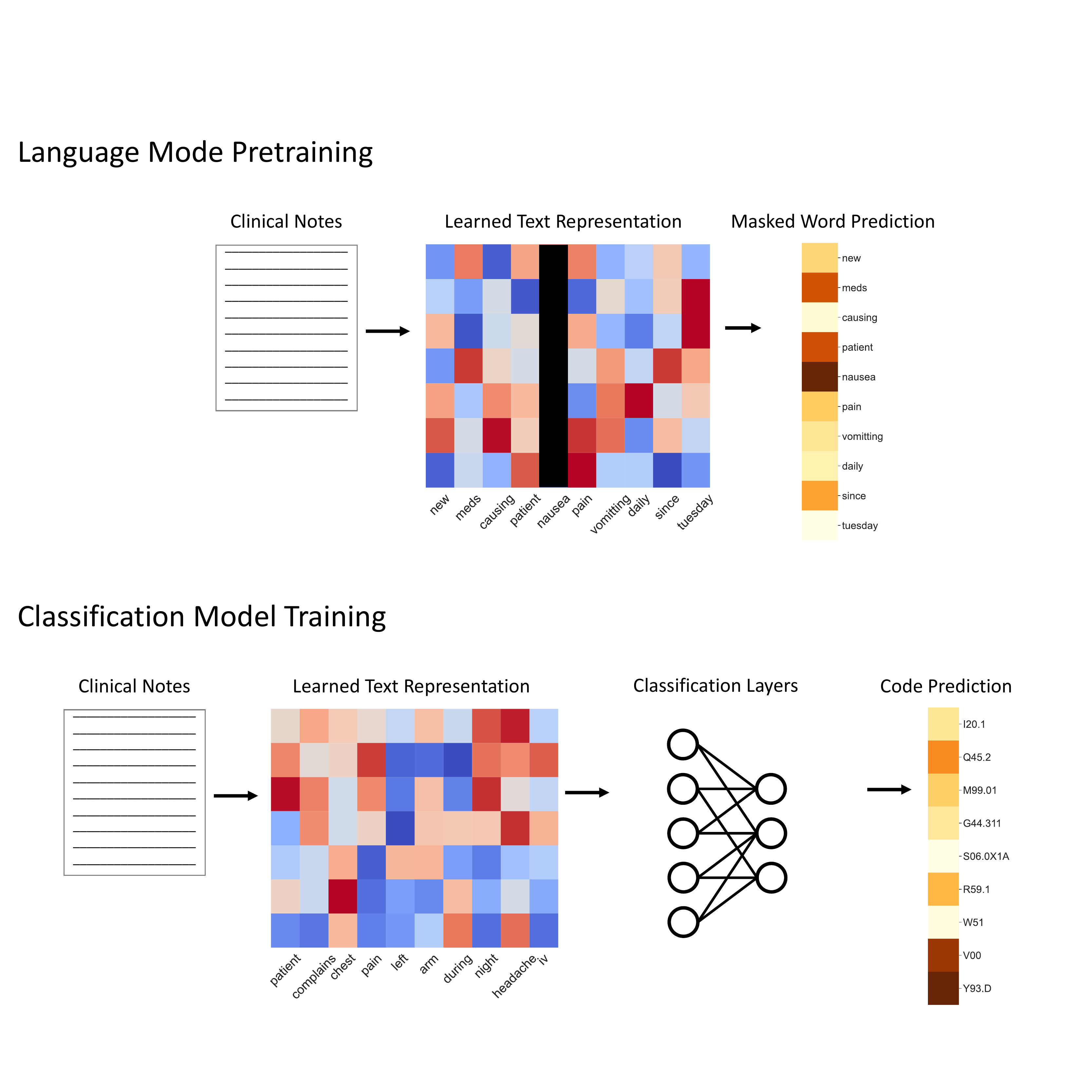}
  \caption{High-level overview of pretraining and fine-tuning approach for clinical text.
  (Top) During pretraining, vector-based representations of tokens are learned via masked language modeling (MLM)
  where 15\% of tokens are masked from the input sequence and the remaining tokens are used to predict missing tokens,
  thus, incorporating word context into the learned representations.
  (Bottom) During fine-tuning, the whole system is trained end-to-end with new labels,
  jointly updating token representations while learning to predict the provided labels.
  We focus on the task of medical condition prediction where the labels are ICD codes representing diagnoses.}
  \label{fig:overview}
\end{figure*}

\subsection{Model Development and Training}
\label{sec:appendix_model_development}

We implemented the deep learning models described using PyTorch \citep{paszke_pytorch_2019} (version 1.9.1).
For all models except the Big Bird LM, we selected the best model hyperparameters using 20 iterations of
Bayesian grid search with version 2.10.0 of the Optuna library \citep{akiba_optuna_2019} and a batch size of 112.
Final hyperparameters were selected according to the best micro-average-precision over 2-fold cross-validation.
Because some conditions are rare, we applied iterative stratified splitting \citep{sechidis_stratification_2011}
to create folds during hyperparameter optimization and when creating train, validation,
and test sets to ensure a roughly uniform distribution of labels across data splits.
Using the best hyperparameter settings, the final models were trained on the entire training set.
We used early stopping based on validation loss for all models,
loading the best checkpoint according to lowest validation loss as the final model.

We used the learning rate finder proposed in \cite{smith_cyclical_2017} to select learning rates for each model and used a starting learning rate of 0.001.
For additional details related to preprocessing and model training, see Supplemental Methods~\ref{sec:data_loading}.

\subsubsection{Language Models}
\label{sec:appendix_model_development_language}

With TF-IDF \citep{sparck_jones_statistical_1988}, we represented each document as a bag of phrase frequencies
using word-level 3 and 4 grams occurring with 0.001 to 0.5 frequency across all documents.

With word2vec \citep{mikolov_efficient_2013} we represented text sequences as embedding vectors concatenated
in order of occurrence into a matrix $X = [x_1; x_2,\dots, x_N]$ where $x_n \in \mathbb{R}^{d_e}$ is the $n^{th}$ embedding
vector and $N$ is the sequence length. To train word2vec on our data, we used a word-level vocabulary consisting of the 50,000 most common words identified across
our pretraining dataset consisting of 5,481,937 medical charts represented as text sequences and truncated to the first 32,768 words.
We trained word2vec embeddings of dimension 128 using the continuous bag of words training objective,
whereby the neural network used to create the embeddings takes
as input a one-hot representation of the five word-level tokens to the left and right of a given token, and
uses this information to predict the given token.
This process was repeated for all words in all input sequences,
and we trained the model for five epochs.  
All other parameters were set to the defaults in version 3.8.3 of the gensim \citep{rehurek2011gensim} library for Python.
We set the embedding layer of the CNN models to trainable, enabling updates to the word representations during model training.
for the supervised classification task.

The parameters of the very long Big Bird LM are shown in Table~\ref{tab:big_bird_params}.
The main aspect of the architecture we modified is the maximum sequence length, which
poses challenges for training, especially data loading.
Consider that each token in each input sequence is represented by a vector and the multiple attention heads
of the LM run the block sparse self-attention operation in parallel, processing the same sequence
with multiple heads at once.  Even without full self-attention, the training process is memory intensive.

We selected hyperparameters similar to the base Big Bird model in \citet{zaheer_big_2021} and report them in Table~\ref{tab:big_bird_params}.
We pretrained the model for the MLM objective and measured validation loss after every 200 steps.
We could fit only one sample at a time on GPUs with 32GB RAM, and so used
a batch size of 32 (there were 32 GPUs total in the cluster) with four gradient accumulation steps for an effective batch size of 128.
Training for 50 steps took approximately 20 minutes, and so we limited training to three full epochs,
after which, decreases in validation loss of 0.01 took over 20,000 steps. 
We expect further pretraining of the LM to lead to better downstream performance and leave this as future work.


When pretraining the Big Bird model with a 4,096 token maximum sequence length on the \optumname dataset,
we trained for the same number of steps, using the same hardware configuration and model hyperparameters
as described above except for the maximum sequence length.  We used the same pretraining dataset but split documents 
into chunks of 4,096 tokens, such that some medical charts were represented as multiple samples.
Samples from medical charts for the same patient were always in the same data split.
For example, if multiple samples from a given patient were in the validation set, 
all samples for that patient were in the validation set, and so forth for all data splits 
to avoid leakage.

\input{tables/big_bird_params.tex}

\subsubsection{Text Classifiers}
\label{sec:appendix_model_development_classification}

Using the TF-IDF representation of a text document, we trained two different classification models for baseline comparison:
logistic regression (LR) with ElasticNet \citep{zou2005regularization} regularization, and a multi-layer FFNN.

For the ElasticNet model, we performed one round of feature selection using the Lasso penalty selected from the range 1 to 1,000,000 over 20 Optuna trials.
We then applied the ElasticNet penalty with L1:L2 ratio selected from the range 0 to 1,
and regularization strength selected from the range 1 to 1,000,000.  For the FFNN, we selected from between 1 and 4 hidden layers
of size between 4 and 192 and dropout rates between hidden layers between 0.05 and 0.5.

In addition to the ElasticNet and FFNN baselines, we fine-tuned several short LMs pretrained on public text data such as Wikipedia on the MIMIC dataset to predict the 50 conditions of interest.
For details on the models and pretraining datasets, please see
the \href{https://huggingface.co/roberta-base}{roberta-base}, \href{https://huggingface.co/emilyalsentzer/Bio_ClinicalBERT}{Bio-ClinicalBERT},
and \href{https://huggingface.co/google/bigbird-roberta-base}{bigbird-roberta-base} (Public Big Bird) model cards from Hugging Face.
Because the BERT models (RoBERTa and Bio-ClinicalBERT) can only represent sequences of 512 tokens at maximum, we tested truncating text to the first 512 tokens and pooling predictions
over many overlapping representations of a full length sequence by taking either the mean or max prediction from each sequence chunk during fine-tuning.
These pooling strategies are based on comments from the BERT author, Jacob Devlin, here\footnote{ \url{https://github.com/google-research/bert/issues/27\#issuecomment-435265194} }.
The code used to fine-tune these models accordingly was modified from this implementation\footnote{ \url{https://github.com/mim-solutions/roberta_for_longer_texts} }.
Bio-ClinicalBERT is pretrained on MIMIC notes, resulting in some leakage when predicting on the MIMIC 50 test set in our experiments.
We take the fact that the clinically pretrained CNN and Big Bird models outperform the fine-tuned Bio-ClinicalBERT models even with leakage as evidence that
pooling over predictions on windowed chunks of the input text is inferior to text-based CNNs and long document LMs implementing sparse self-attention which can represent full length sequences.
All BERT models were fine-tuned using Hugging Face Transformers 4.10.3 and PyTorch 1.9.1 with a maximum learning rate of 0.00001, a linear warmup for 1000 steps, and the AdamW optimizer with 0.01 weight decay.
All BERT models used early stopping on validation loss, training until 20 epochs of no improvement.
Effective batch size (with gradient accumulation) for all BERT training runs was 128.
For pooling, windows of 510 tokens were created with 128 token overlaps.
The custom aggregation function for Bio-ClinicalBERT was implemented as described in equation 4 of \citet{huang_clinicalbert_2020} using $c=2$ as described in the paper.

Because the very long, clinically pretrained Big Bird model used the MLM objective, we followed the practice outlined in the Big Bird paper
of warming up the learning rate as we fine-tuned.  We used 2,000 warm-up steps and an inverse square root learning rate schedule
to linearly increase the learning rate over the warm-up steps until we hit 0.00005, then square root decayed the learning rate
over subsequent steps when fine-tuning on the \optumname dataset.
We used an effective batch size of 128 for all fine-tuning experiments (32 samples per GPU with four gradient accumulation steps)
and the AdamW \citep{loshchilov_decoupled_2019} optimizer with 0.001 weight decay.

Because sequences in the MIMIC dataset are significantly shorter than the \optumname dataset, we were able to fine-tune the
off-the-shelf and clinically pretrained 4,096 maximum sequence length Big Bird models on the MIMIC dataset to predict medical conditions from the text.
On this dataset, we tested both the Big Bird models we pretrained on the \optumname dataset and a Big Bird model
pretrained on the Books, CC-News, Stories and Wikipedia datasets with a maximum sequence length of 4,096 tokens \citep{zaheer2021big}.
We fine-tuned the 4,096 sequence length models with 0.01 weight decay in AdamW, a
linear learning rate schedule, and 1000 warm-up steps using transformers version 4.10.3 \citep{wolf-etal-2020-transformers}.

\subsection{Data Loading}
\label{sec:data_loading}

All data preprocessing happened on a single CPU machine with 672 GB RAM and 96 cores.
We pre-tokenized the data in batches so that when data was loaded on the fly,
the model did not need to wait for any transformations to be applied.

To train the ElasticNet \citep{zou2005regularization} LR, FFNN, and CNN models in our experiments, we applied a straightforward data loading procedure.
This consisted of splitting batches of 112 training samples onto four GPUs one at a time by first loading multiple batches from disk,
splitting one batch onto the four GPUs, moving to the next batch,
and loading the next set of multiple batches once all batches in memory were exhausted.

To pretrain the Big Bird long document LM on 5,481,937 sequences of 32,768 subwords, we created input tensors of tokenized and masked text prior to training.
Even before moving tensors to the GPU, we were limited by the size of tensor we could fit into RAM.
As such, training tensors were chunked and loaded into memory one at a time.
We broke the charts into chunks of 6,400 samples each and one chunk with less than 6,400 samples.
We considered a full epoch to be one pass through all chunks.
We considered one training round to be a pass through 40 chunks followed by evaluation on 64,000 samples to compute validation loss, using this for early stopping.

To accommodate a validation set of arbitrary size, validation set sequences were saved as individual samples and loaded on different GPUs during validation epochs
according to a mapping used by the distributed sampler.
The mapping pointed a randomly sampled index to a file for that sample, similar to how image classification models often load and train on individual images.

Initial tests of saving 64,000 torch arrays with pickle were very slow, about 500 files per hour.
Saving 64,000 numpy arrays with "np.save" was much faster, about 30,000 files per minute.
A Stack Overflow post\footnote{ \url{https://stackoverflow.com/questions/9619199/best-way-to-preserve-numpy-arrays-on-disk} } provides
a nice comparison of array I/O performance for many file formats.
File loading was also significantly faster with numpy. Reading 1000 files took about 30 seconds,
whereas it took about 30 seconds to read just one pickled torch array during initial tests.

\subsection{Distributed Training of Very Long Big Bird LM}
\label{sec:distributed_long_lm_training}

We trained on a cluster of four Standard\_ND40rs\_v2 VMs in Azure with eight GPUs, 40 cores, 672 GB CPU RAM, and 32 GB GPU RAM each.
With batches of one record per GPU (32 total), and four gradient accumulation steps (effective batch size of 128),
training on one chunk took 21.3 minutes on average.
Predicting on all validation samples took 216.3 minutes on average.
Thus, the total time for one round of training was $40 \cdot 21.3 + 216.3 = 1,068.3$ minutes and a full epoch with 21.4 training rounds took 15.7 days.
We trained for three epochs based on a fixed budget, but in future work, aim to pretrain for
longer and focus on ways to improve pretraining efficiency.




\begin{algorithm2e*}[h]
  \caption{Masked Sampling Significance Test}\label{alg:masking_sig}
  \DontPrintSemicolon
  \KwData{$\hat{y} \in \mathbb{R}^{L}$, maskedSampleProbs $\in \mathbb{R}^{N \times L}$, maskIndices $\in \{0,1\}^{N \times (S_{i}/B)}$}
  \KwResult{p-values $\in \mathbb{R}^L$}
  \SetKwFunction{FSum}{MaskedSamplingSignificanceTest}
  \SetKwProg{Fn}{Function}{:}{}
  \SetKw{KwBy}{by}
  \Fn{\FSum{$\hat{y}$, maskedSampleProbs, maskIndices, blockIndex, sizeBootstrapSample, numBootstrapIters}}{
    p-values $\gets$ [ ]\;
    \For{$l=1$ \KwTo $L$}{
      $\overline{\Delta \hat{y}_{l}} \gets \frac{1}{N} \sum_{n=1}^{N} \Delta \hat{y}_{n,l}$\;
      blockIndices = \{i if $\textnormal{maskIndices}_{i,blockIndex}$ = 1\}\;
      $\Delta \hat{y}_{l,blockIndex} \gets \hat{y}_{l} - \textnormal{maskedSampleProbs}_{\{blockIndices\},l}$\;
      bootstrapScores $\gets$ [ ]\;
      \For{$b=1$ \KwTo $numBootstrapIters$}{
        randomScores $\gets$ sampleWithReplacement($\Delta \hat{y}_{l,blockIndex}$, size=sizeBootstrapSample)\;
        bootstrapScore $\gets \frac{1}{sizeBootstrapSample} \sum randomScores$\;
        bootstrapScores.append(bootstrapScore)\;
      }
      p-value $\gets \frac{\textnormal{count}(bootstrapScores > \overline{\Delta \hat{y}_{l}})}{numBootstrapIters}$\;
      p-values.append(p-value)
    }
    \KwRet p-values\;
  }
  \end{algorithm2e*}

\subsection{Blind Experiment Sampling}
\label{sec:blind_experiment_sampling}

To measure the performance of MSP against the SOC algorithm,
we had two clinical reviewers annotate the informativeness of
text blocks deemed important by MSP and SOC.
To run these algorithms, we randomly sampled 40 discharge summaries from the MIMIC 50 test set,
ran MSP on the first 20, SOC on the second 20,
and took the top $K=5$ most important blocks for each positive label (ICD-9 code) from each.
We also randomly sampled text blocks from the discharge summaries, the same number of which
were identified via MSP.
This resulted in 1,850 line items having valid ICD-9 descriptions.
To limit clinical review time to about six hours, estimating 200 lines items per hour,
we randomly sampled 400 ICD-9 label-document combinations for each of the three algorithms from the 1,850 line items.
This resulted in a total of 1,200 line items which we provided to each clinical reviewer.

\section{Supplemental Results}
\label{sec:supplemental_results}

\subsection{Additional Performance Plots}
\label{sec:additional_performance_plots}

\begin{figure*}
  \centering
  \subfigure[]{\includegraphics[width=0.495\textwidth]{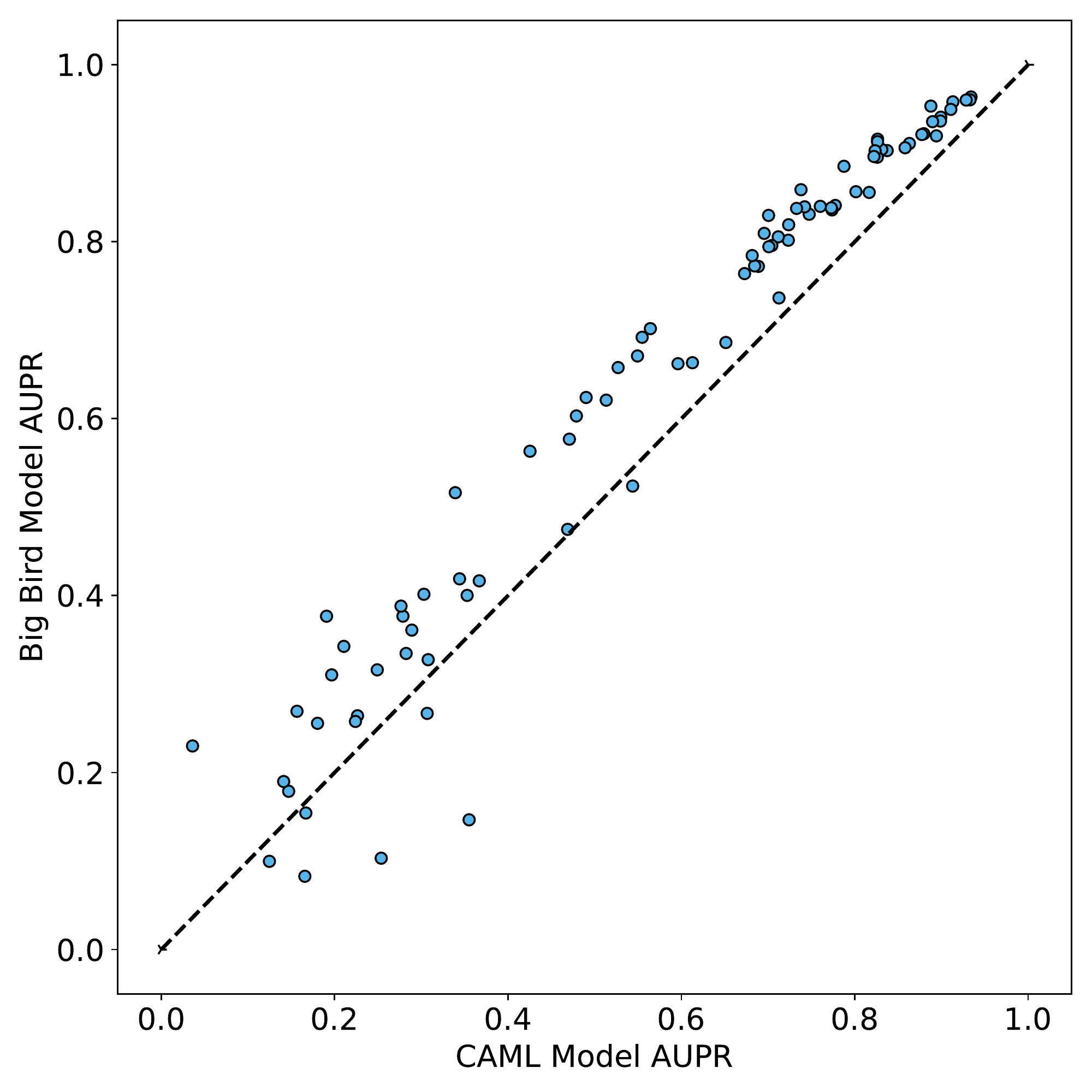}}
  \subfigure[]{\includegraphics[width=0.495\textwidth]{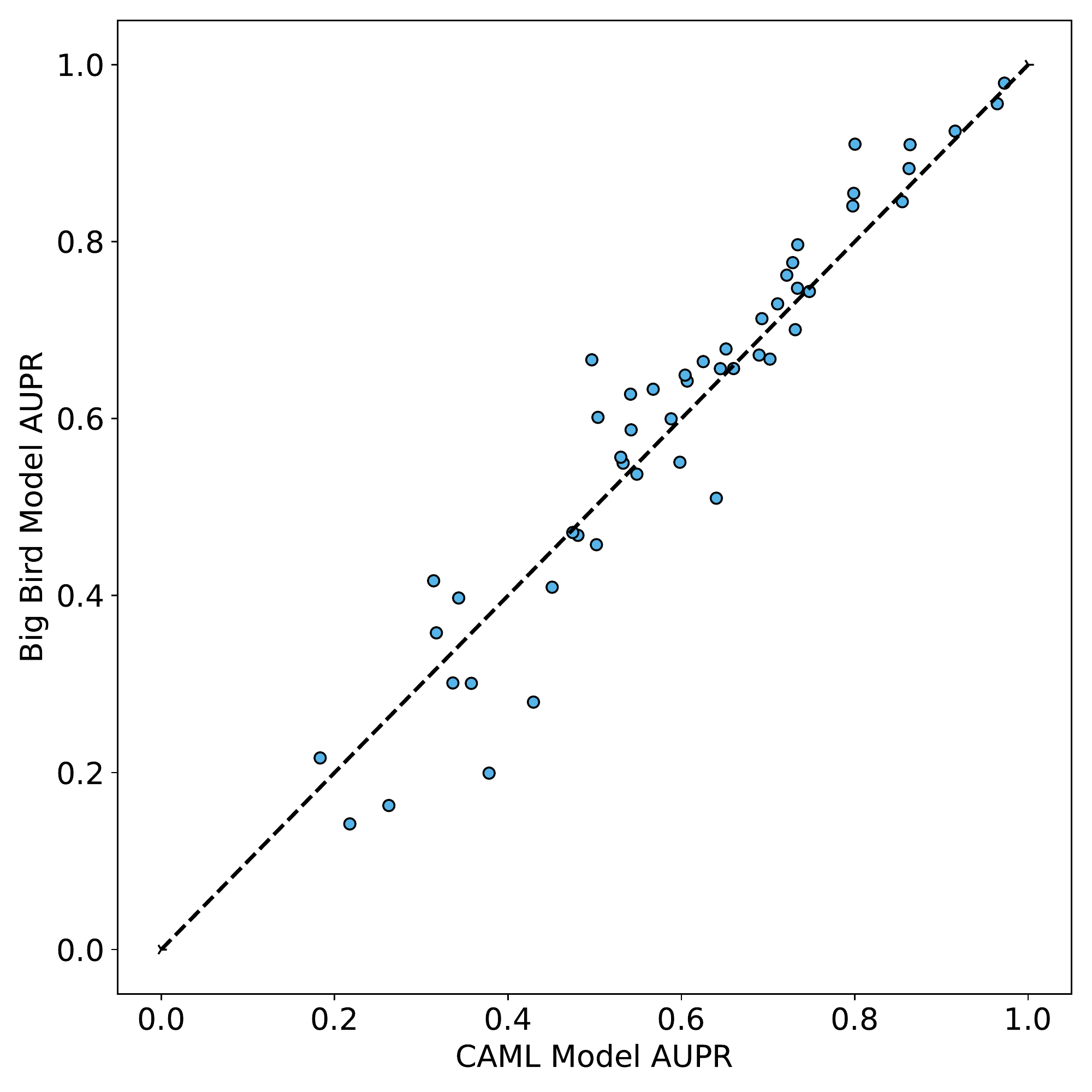}}
  \vskip\baselineskip
  \subfigure[]{\includegraphics[width=0.495\textwidth]{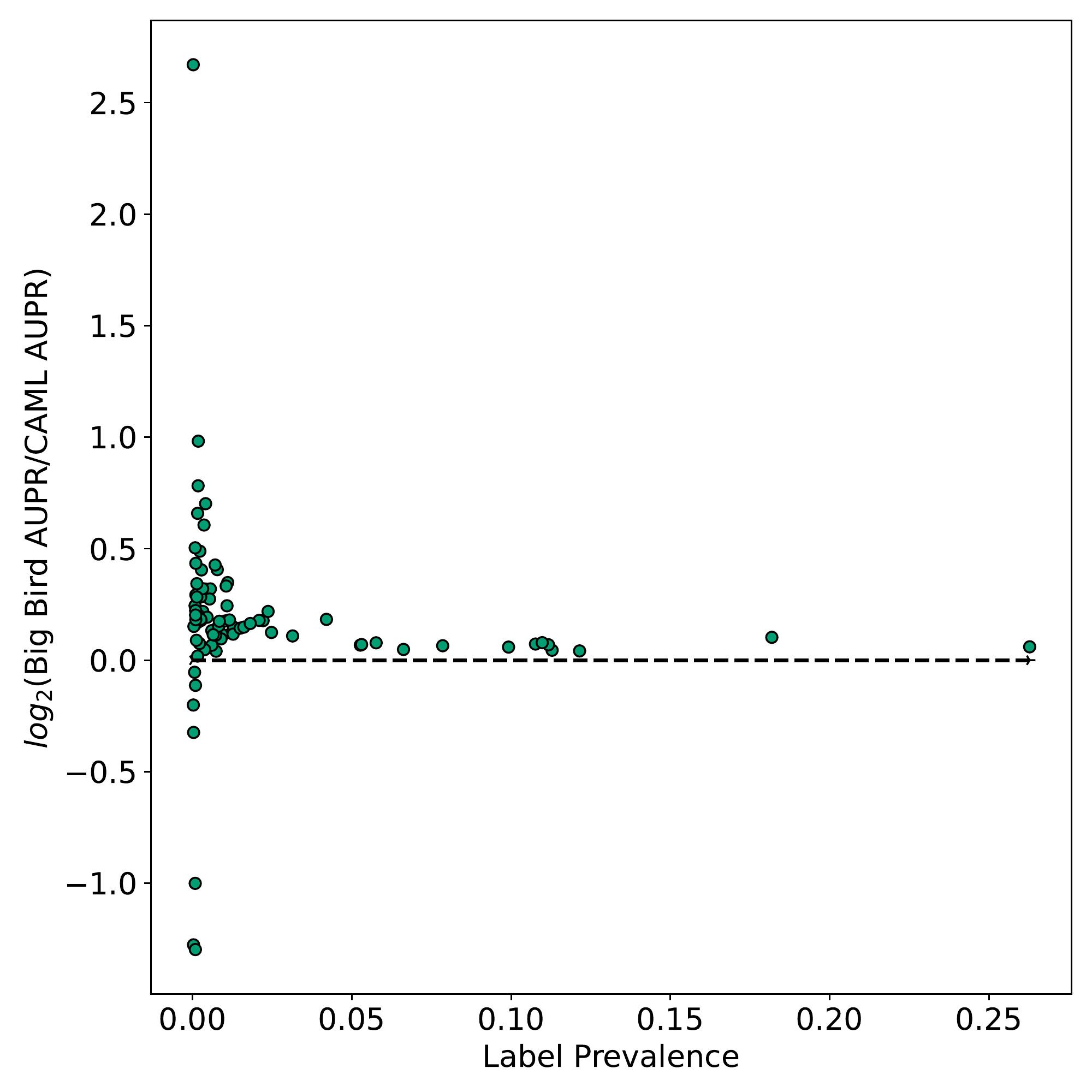}}
  \subfigure[]{\includegraphics[width=0.495\textwidth]{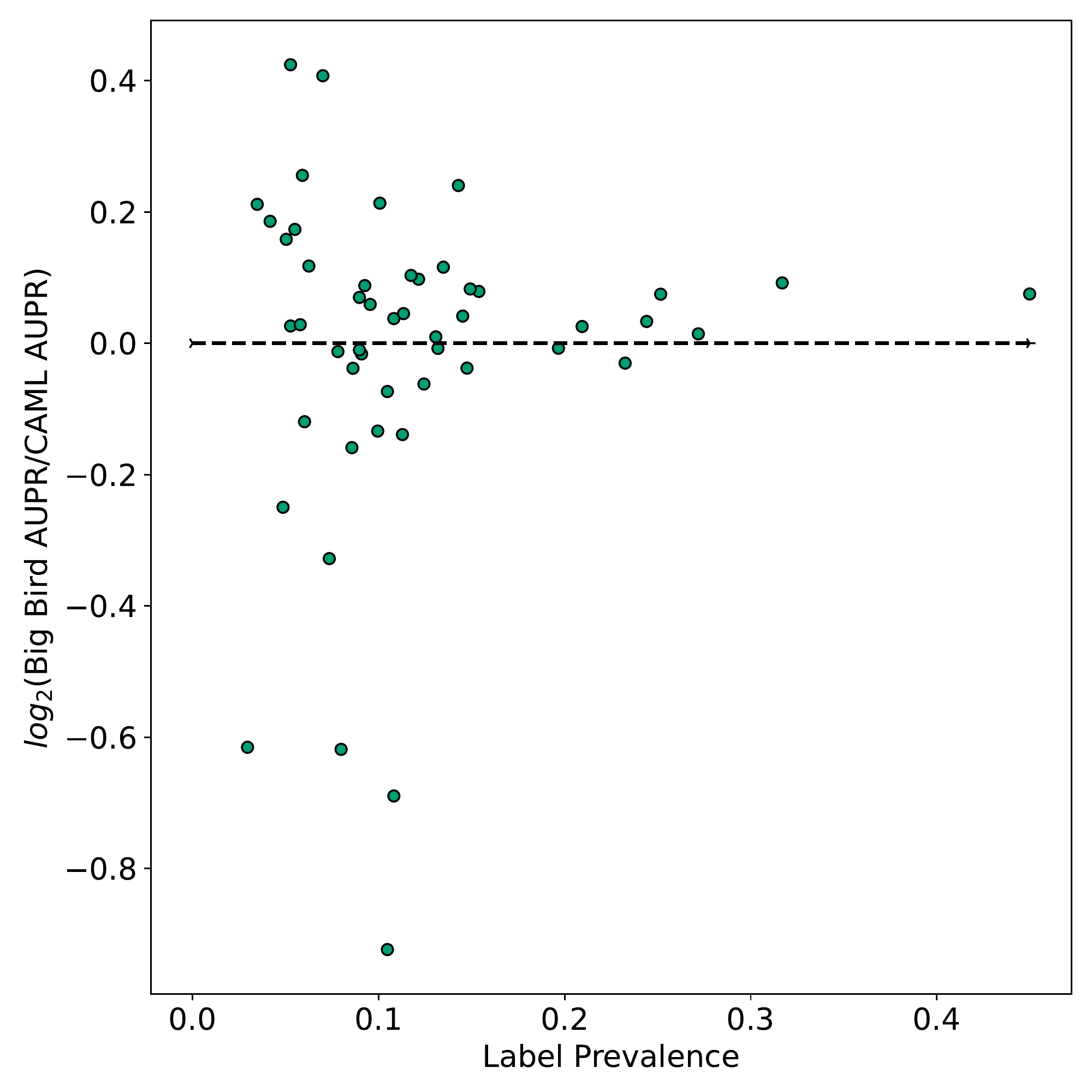}}
  \caption{Area under the precision-recall (AUPR) curve per code label for the CAML model (x-axis) compared to the 32,768 max sequence length Big Bird model (y-axis) for the \optumname dataset using the largest training set consisting of 640,000 charts (a) and MIMIC-III data (b).
  In (a) and (b), the dashed black line $y=x$ denotes equal performance between the two models.
  For each label, we also compare the $\textnormal{log}_2$-scaled ratio of the Big Bird AUPR to the CAML AUPR as a function of label prevalence in both the \optumname dataset (c) and MIMIC-III dataset (d).
  In (c) and (d), the dashed black line $y=\textnormal{log}_2(1)=0$ denotes equal performance between the two models, with points above the line showing where Big Bird performs better and points below the line showing where CAML performs better.}
  \label{fig:best_model_comparison}
\end{figure*}

\subsubsection{\optumname Dataset}

In Supplemental Figure~\ref{fig:all_model_comparison} we show micro- and macro-averaged model performance using area under the ROC (AUROC) curve, AUPR, and the F1 metric.
Of the models evaluated, we found that models utilizing the TF-IDF representations had the poorest performance across all metrics.
We expected this to be the case, as the TF-IDF text representation fails to account for word order.
The CAML model with pretrained word2vec embeddings outperformed the TF-IDF models, but was worse on average than the very long Big Bird model.
The Big Bird model outperforms the CAML model by a statistically significant margin in every classification metric.
Because all layers of the Big Bird LM are pretrained on the text of medical charts, we expect the model carries more information than the individual word vectors used in the CAML model.
Additionally, while the attention mechanism in the CAML model enables tokens near each other in an input sequence to attend to each other,
tokens in the Big Bird model can attend to each other across distant pieces of the input sequence via global tokens and random connections.
We explore this further in Supplemental Results~\ref{sec:integrating_distant_contextual_information}.

\begin{figure*}[hb]
  \centering
  \includegraphics[width=5.5in]{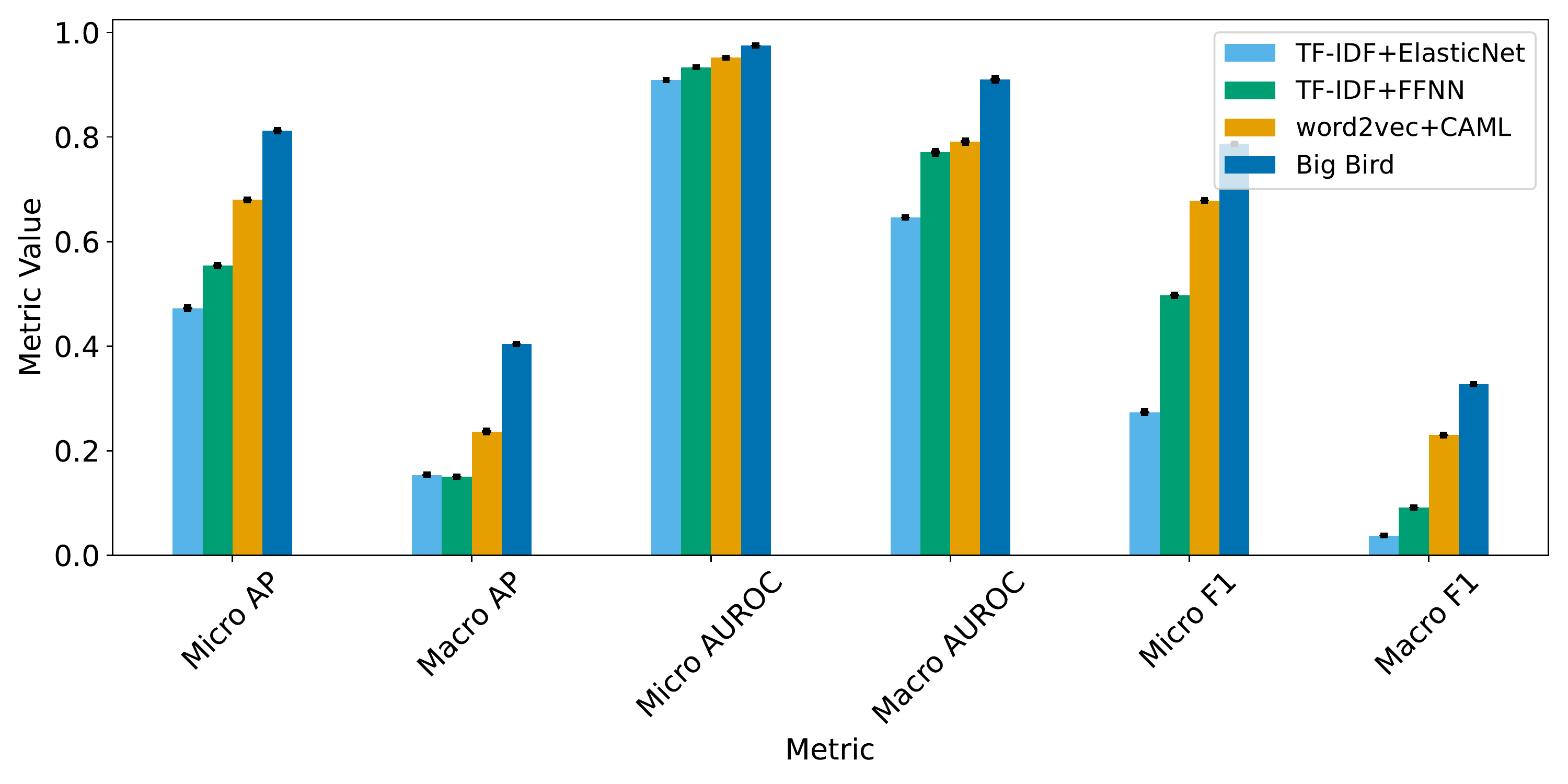}
  \caption{Performance metric comparison for the \optumname dataset using the largest training set consisting of 640,000 charts for all models evaluated.
  Micro-averaging of classification metrics takes label imbalance into account by taking a weighted average of the performance metric.
  Macro-averaging takes an unweighted mean of the performance metric across all labels.
  AP: average precision; AUROC: area under the ROC curve; F1: harmonic mean of precision (PPV) and recall (sensitivity).}
  \label{fig:all_model_comparison}
\end{figure*}

When comparing the performance of each model for a particular medical code, it is important to include the label prevalence for context.
Whereas the AUROC for a random model is 0.5 regardless of the label prevalence, for AUPR the random model performance is equal to the label prevalence.

In Supplemental Figure~\ref{fig:pr_barplot}a we show the AUPR for the CAML and Big Bird models for the top 10 medical conditions where the difference between the Big Bird AUPR and the CAML AUPR was the greatest.
As a baseline for each condition we also show the prevalence.
The improvements of the very long Big Bird model over the CAML model appear largest when the prevalence is low.

\begin{figure*}
  \centering
  \subfigure[]{\includegraphics[width=0.495\textwidth]{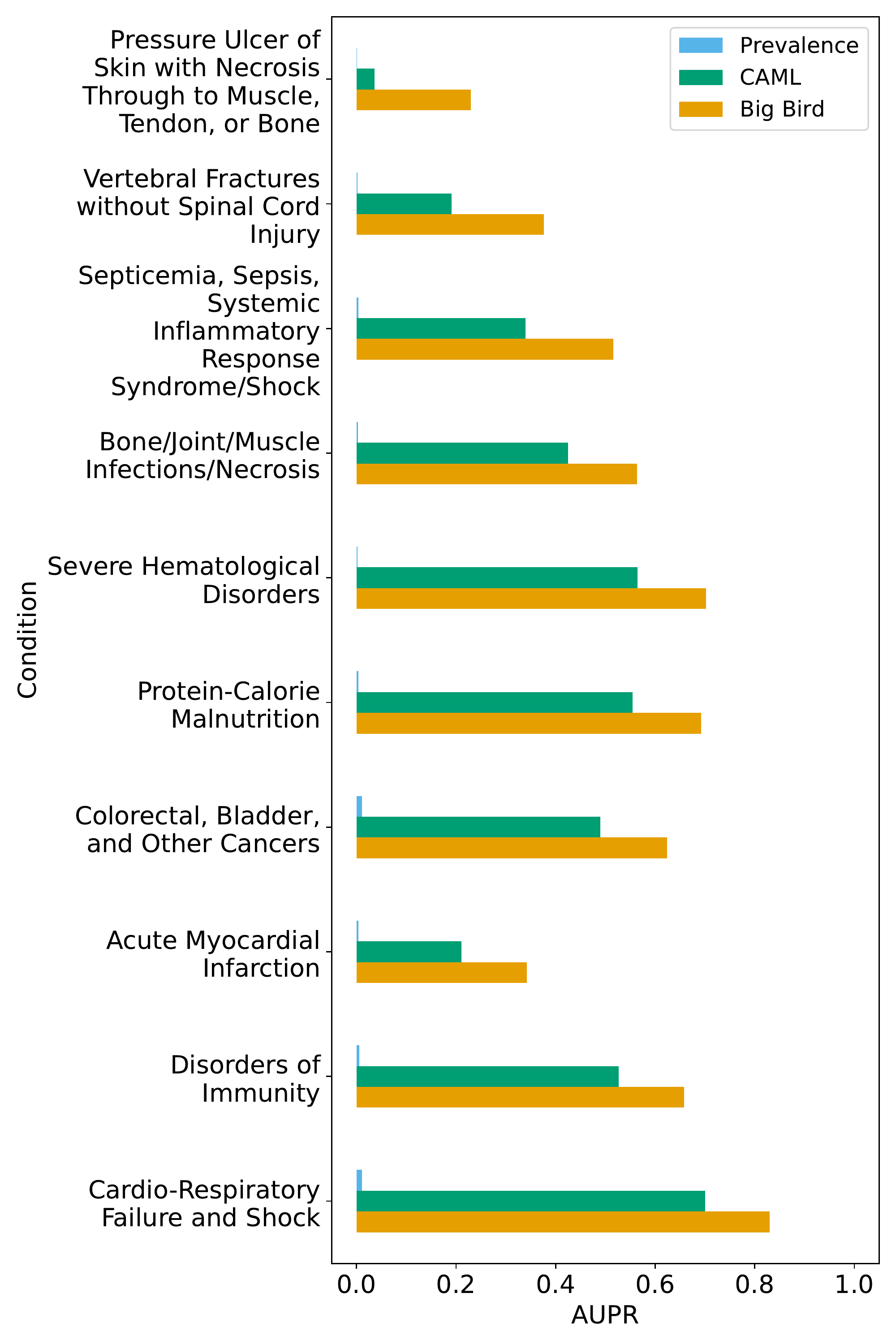}}
  \subfigure[]{\includegraphics[width=0.495\textwidth]{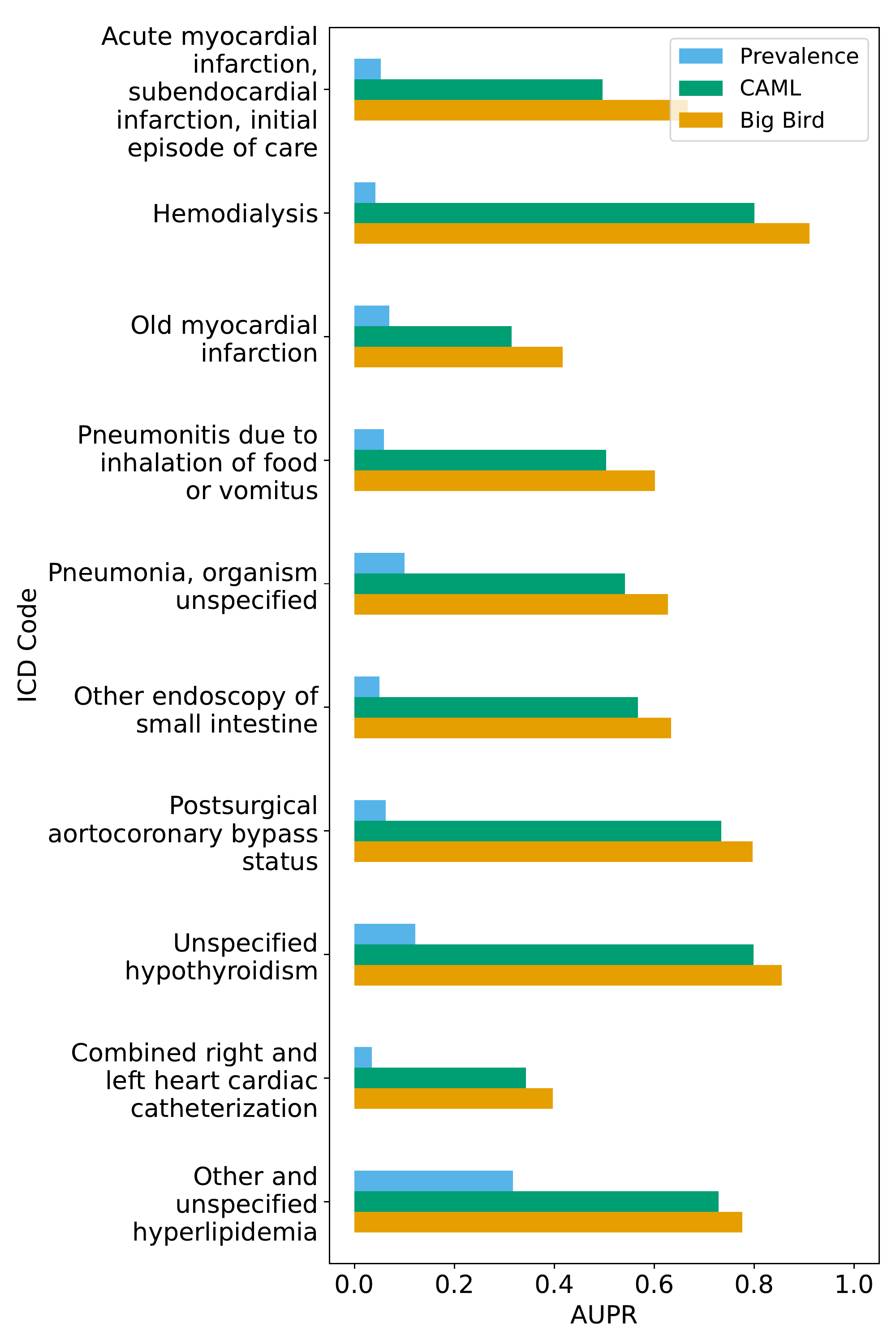}}
  \caption{Comparison of AUPR values for the top 10 conditions where the difference between the Big Bird model and the CAML model performance was the largest (i.e. Big Bird improved over the CAML model) for (a) the \optumname testing set, using models trained with 640,000 charts, and (b) the MIMIC-III testing set. The prevalence is the fraction of documents with an occurrence of the medical condition.}
  \label{fig:pr_barplot}
\end{figure*}

\subsubsection{MIMIC Dataset}

Figure~\ref{fig:pr_barplot}b shows
the AUPR for the CAML and Big Bird models for the top 10 ICD codes where the difference between the Big Bird AUPR and the CAML AUPR was the greatest.

\subsection{Performance Tables}
\label{sec:performance_tables}

The main text of the paper refers to both average and per label performance metrics for both datasets.
Figure~\ref{fig:all_model_comparison} compares the test set performance of all models trained using the largest
training set from the \optumname dataset.
Average performance over all medical conditions for the various training sets from the \optumname dataset can be found in
Table~\ref{tab:12800_perf_transpose}, Table~\ref{tab:64000_perf_transpose},
Table~\ref{tab:128000_perf_transpose}, and Table~\ref{tab:640000_perf_transpose}.
Average performance over all medical conditions for the MIMIC dataset can be found in Table~\ref{tab:8067_perf_transpose}.
Per label metrics for both datasets can be found in
Table~\ref{tab:optumcharts85_performance_compare} and Table~\ref{tab:mimic50_performance_compare} respectively.

\input{tables/12800_perf_transpose.tex}
\input{tables/64000_perf_transpose.tex}
\input{tables/128000_perf_transpose.tex}
\input{tables/640000_perf_transpose.tex}
\input{tables/8067_perf_transpose.tex}
\input{tables/bert_mimic_results.tex}

\newpage
\input{tables/optumcharts85_performance_compare.tex}
\newpage
\input{tables/mimic50_performance_compare.tex}

\subsection{Effect of Training Set Size}
\label{sec:effect_of_training_set_size}

\begin{figure*}[t]
  \centering
  \includegraphics[width=5.5in]{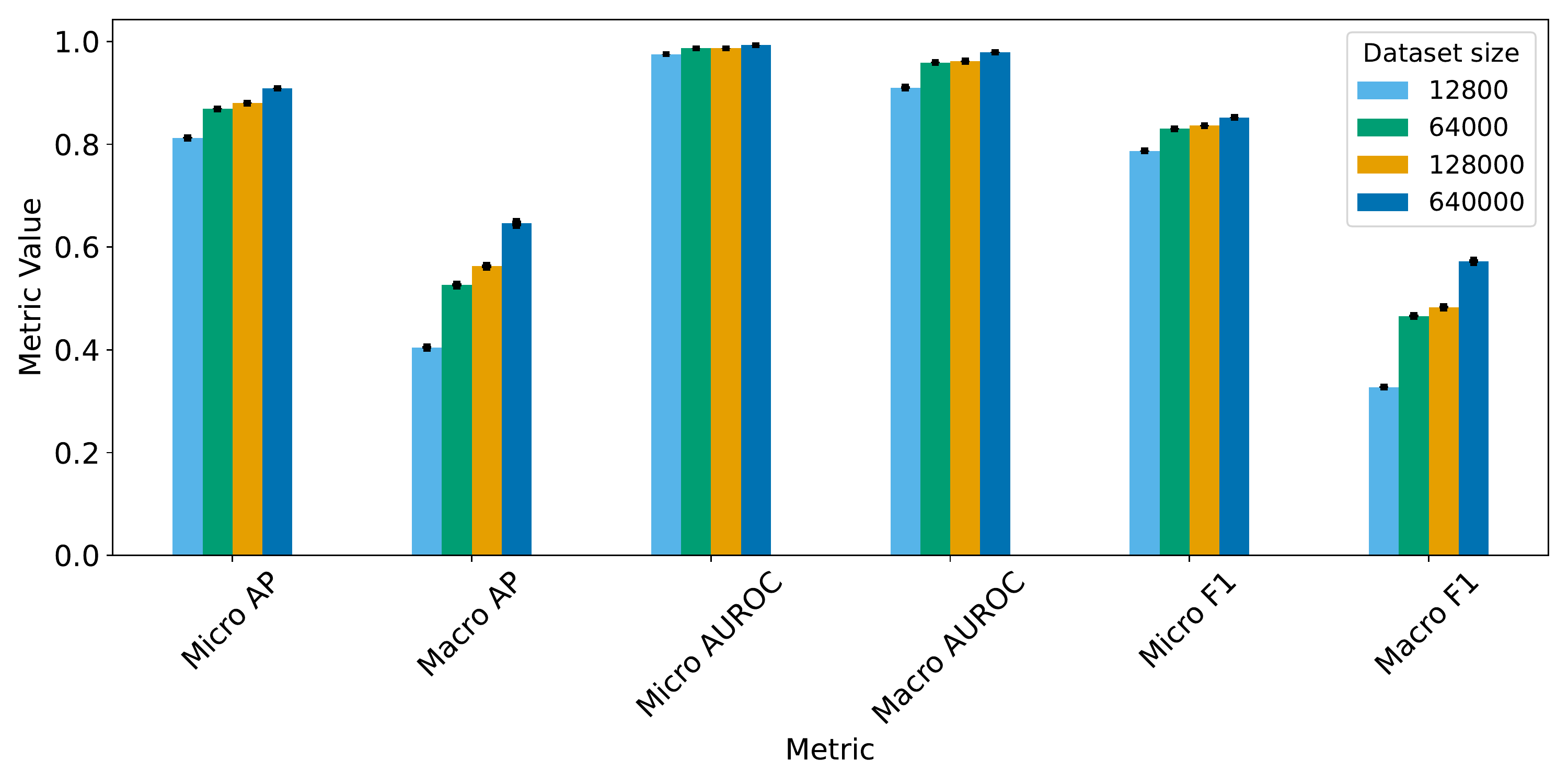}
  \caption{Big Bird model performance metrics as a function of the number of samples used for model training. Error bars with the 95\% confidence interval ($1.96 \times$ standard deviation) are shown in black.}
  \label{fig:performance_vs_sample_size}
\end{figure*}

Traditional supervised models with a large number of parameters require a large number of examples to achieve good performance.
However, transfer learning with pretrained LMs has demonstrated that for some datasets, only a small amount of labeled data is required
to achieve good performance.
We measured the performance of the pretrained, very long Big Bird LM on fine-tuning datasets of varying size.
To do this we used a random sample of 5,481,937 unlabeled charts for pretraining word2vec embeddings and the Big Bird LM.
We then created random samples of 12,800, 64,000, 128,000, and 640,000 labeled charts for training, 64,000 labeled charts for validation, and 187,953 labeled charts for testing.
Figure~\ref{fig:performance_vs_sample_size} shows the performance of the pretrained Big Bird LM relative to the CAML model with pretrained word embeddings
as we increase the size of the training dataset.
As is expected with data-hungry ML models, performance increased with the number of training samples.
Compared with the AUROC, the average precision (AP) saw a bigger increase in performance with sample size.
Also of note is that the macro-averaged (unweighted) metrics saw the biggest improvement compared to the micro-averaged (prevalence-weighted) metrics,
which suggests that increasing the sample size has a larger benefit for the labels that occur less frequently.
Across dataset sizes, we observed a consistent 5\% absolute improvement in micro-average-precision over the CAML model
using the clinically pretrained, very long Big Bird LM, making a strong case for applying this method to very long documents of varying sample sizes.

\subsection{Masked Sampling Procedure Additional Details}
\label{sec:masked_sampling_procedure_additional_details}

\input{tables/masking_experiment}

Table~\ref{tab:masking_experiment} depicts the proportion of informative text blocks
identified using MSP at various values of $P$ and randomly selected text blocks where informative text blocks
were annotated by a single clinical reviewer.
In these experiments, we set the number of iterations, $N$,
such that the expected number of times a given text block is masked when computing importance is equal to 1000.
For example, when the masking probability, $P$ is 0.1, we set $N=10,000$.
For each sampled discharge summary, we selected the $K=5$ most important text blocks for each positive label (ICD-9 code).
For each masking probability tested, and for the random set of blocks, the clinician received 117 samples.
We chose 117 by randomly sampling at least five discharge summaries for each masking probability, $P$,
and taking the minimum number of ICD-9 label and discharge summary combinations associated with each $P$,
such that the number of samples provided to the clinician was equal for each masking probability.
Of the 117 text blocks deemed important by the random masking procedure for the best value of $P$,
35 (29.9\%) were considered relevant to the diagnosis according to clinical review.

All values of $P$ except for $P=0.9$ are significant with $\alpha=0.05$ and remain significant after Bonferroni correction.
These results suggest that MSP indeed identifies text blocks relevant to the predicted medical condition labels.
Based on the proportion of informative blocks for each $P$, we hypothesize that lower values of $P$ better isolate the effects of
individual blocks than higher values of $P$ in which most blocks are masked.
The proportion of clinically informative text blocks for all masking probabilities is shown in Table~\ref{tab:masking_experiment}.

\subsubsection{Mean Reciprocal Ranking at K}
\label{sec:mrr_at_k}

\begin{figure*}[h]
  \centering
  \includegraphics{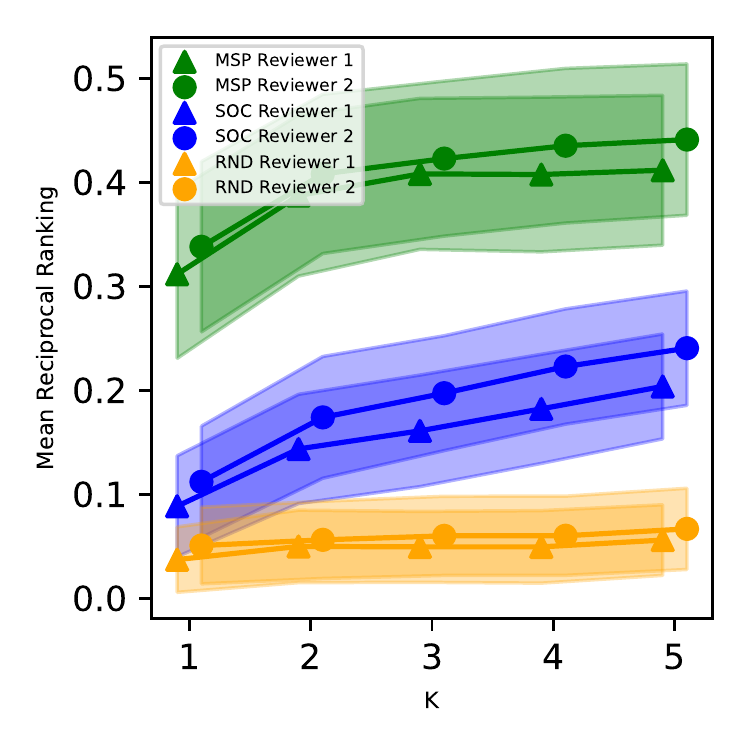}
  \caption{Mean reciprocal ranking (an information retrieval metric) for the top $K$ text blocks surfaced by MSP, SOC, 
  and the random algorithm (RND) according to each reviewer for each document-label pair with 95\% confidence intervals computed using 1000 bootstrap iterations. 
  Mean reciprocal ranking averages performance of each algorithm across the reciprocal of the rank of the most informative text block surfaced among the top $K$ text blocks
  for each document-label pair.  If no blocks surfaced were important for a given document-label pair, the value for that example is $0$.  This metric is valuable 
  in that it privileges algorithms that assign a high rank to informative text blocks.}
  \label{fig:mrrs_at_k}
\end{figure*}

See Figure~\ref{fig:mrrs_at_k}.

\subsubsection{Runtime as a Function of Document Length}

\input{tables/algo_runtimes_len.tex}

LM inference time grows with the length of the document.  As such, the runtime for both MSP and SOC grows with document length,
however, SOC requires additional sampling iterations to compute the importance of each new phrase in a document as document length grows.
Table~\ref{tab:algo_runtimes_len} depicts the change in runtime averaged over 20 trials for various fixed document lengths for the MSP and SOC algorithms.
Note that even at a modest document length of 1000 tokens, identifying the important text blocks in a single document with SOC takes over an hour.

\subsubsection{Integrating Distant Contextual Information}
\label{sec:integrating_distant_contextual_information}

We repeated MSP for pairs of text blocks by identifying which pairs of text blocks have the largest impact on the probability of each label.
For this analysis, we focused on the case where $P=0.1$ and $N=10,000$, such that the expected number of times a given pair of blocks is masked in the same iteration is 100.
For these pairs, we computed the distance between the start of each text block in the pair to understand whether the long document LM is incorporating information from distant
parts of each document in its predictions.
In general, this procedure can be used to identify combinations of many text blocks, and is flexible to different definitions of a block.
We run experiments with $B=10$, identifying important blocks of 10 subword tokens, but blocks could be defined by splitting on punctuation or even entire paragraphs.

\begin{figure*}[th]
  \centering
  \includegraphics[width=5.0in]{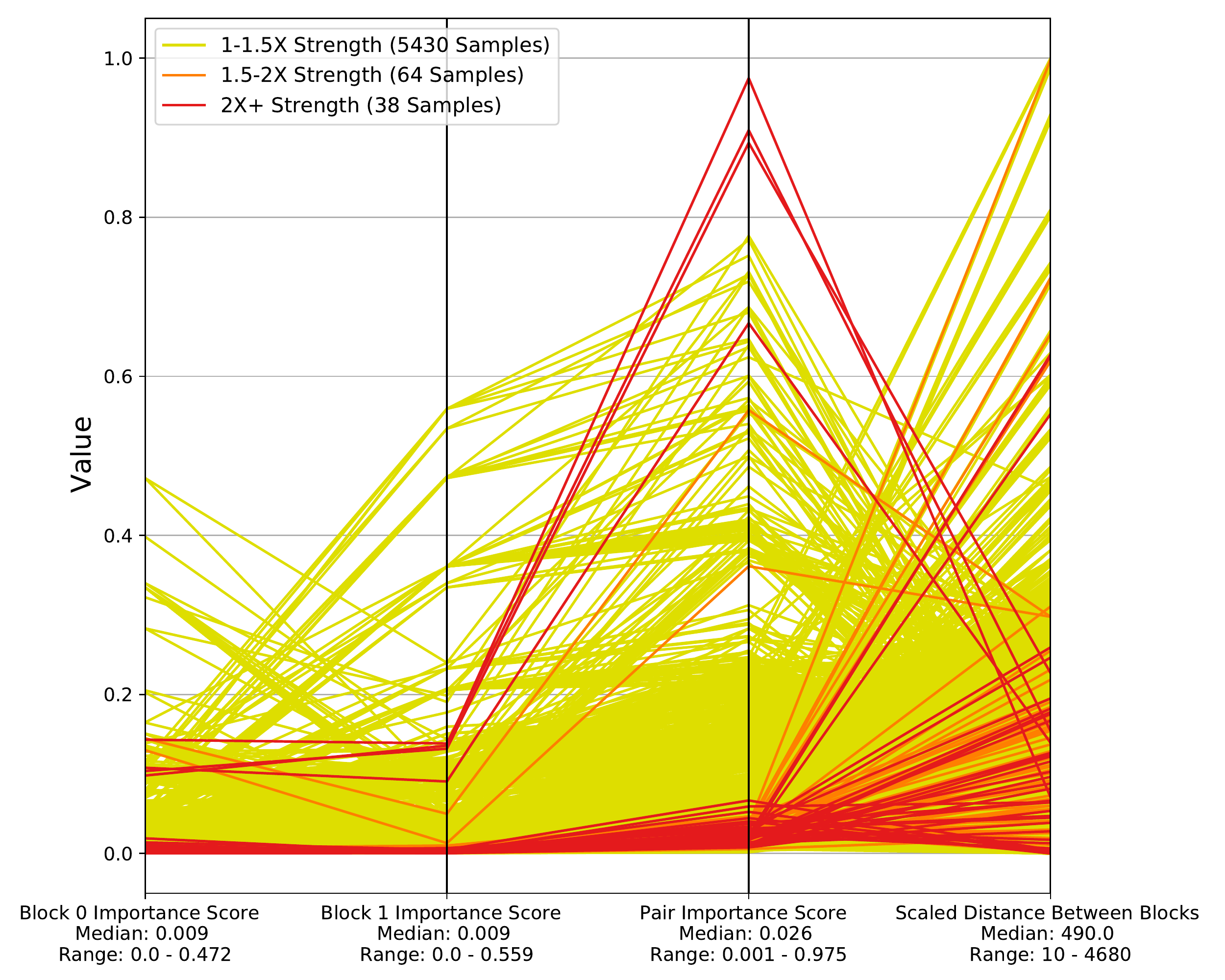}
  \caption{Parallel coordinate plot for block pairs from 15 randomly sampled discharge summaries.
  Interaction strength is computed by comparing the importance of the pair to the sum of the individual importance scores
  for each text block in the pair.}
  \label{fig:interactions}
\end{figure*}

We are interested in block pairs for which the importance score of the pair is greater than the
sum of the importance scores of each block in the pair and consider these cases interactions.
Such interactions would indicate the model recognizes the joint influence on label predictions of text snippets in pairs
beyond the individual contributions of blocks in a pair.
Figure~\ref{fig:interactions} shows the interactions for all block pairs with positive importance scores
for 15 randomly sampled discharge summaries from the MIMIC test set along with the relative distance between blocks in each pair
and the relative strength of the interactions.
Figure~\ref{fig:interactions} illustrates that there are many block pairs for which the combined importance score of pairs
is relatively higher than each block in the pair while the distance between blocks is great, often 100s of tokens (median distance of 490 tokens).
In the Big Bird model,
local attention is applied over windows of 64 tokens (from Supplemental Table~\ref{tab:big_bird_params}), suggesting the model,
through the use of global and random attention across 12 layers, is integrating information
from distant locations within the discharge summaries to predict the ICD-9 labels assigned to each summary.
In this way we demonstrate that not only can MSP be used to identify clinically informative text blocks
used by the long LM to make predictions, it can also uncover pieces of information which, though distant within the document,
in combination, influence the predicted probabilities of ICD labels.

\end{document}

%% file: tables/masked_text_results.tex
\begin{table*}[t]
    \centering 
    \caption{Text blocks deemed informative during clinical review in the order they are mentioned in the main text of the paper.
    We test the null hypothesis that text blocks identified as important by MSP
    are no more important to predicting a diagnosis label than randomly sampled blocks.}
    \begin{tabular}{p{3cm}p{7.5cm}p{1.5cm}l}
        \toprule
        \textbf{Diagnosis} &  \textbf{Text Block} & \textbf{$\Delta P(code)$} & \textbf{p-value}\\
        \midrule
        \midrule
        pneumonia, organism unspecified & pneumonia patient being discharged o n maximal copd regimen including & 0.407 & $<0.001$\\
        \midrule
        subendocardial infarction, initial episode of care & lovenox bridge nstemi o n admission the patient had elevated & 0.420 & $<0.001$\\
        \midrule
        atrial fibrillation & consulted amiodarone was held rhythm slowly began to recover she & 0.186 & $<0.001$\\
        \midrule
        aortocoronary bypass status & al likely improve as pna improves s p cabg complicated & 0.207 & $<0.001$\\
        \midrule
        acute respiratory failure & albuterol and ipra prn his acidosis slowly improved as did & 0.282 & $<0.001$\\
    \bottomrule
    \end{tabular}
    \label{tab:masked_text_results} 
\end{table*}

%% file: tables/msp_vs_soc_vs_rand.tex
\begin{table*}[t]
    \centering
    \caption{Number and proportion of informative text blocks (IBs) identified during blind clinical review.
    We compared MSP at $P=0.1$ to SOC and a random algorithm (RND).  400 text blocks were provided from each algorithm
    to two clinical reviewers who worked independently to score text blocks as informative or uninformative.
    We report p-values from two-tailed, two-sample T-tests without assuming equal variances, comparing the proportion of IBs identified by
    MSP vs RND, MSP vs SOC, SOC vs RND. All tests are significant with $\alpha=0.05$ and remain significant after Bonferroni correction.
    The inter-annotator agreement ratio was 0.96 with Cohen’s Kappa 0.78.}
    \begin{tabular}{lllllll}
    \toprule
      & \multicolumn{3}{l}{\textbf{Reviewer 1}} & \multicolumn{3}{l}{\textbf{Reviewer 2}} \\
      \midrule 
      \textbf{Algo} & \textbf{IBs} &  \textbf{RND}  & \textbf{SOC} & \textbf{IBs} &  \textbf{RND}  & \textbf{SOC} \\
      \midrule
      \midrule
      \multicolumn{1}{l|}{RND} & 7 (1.8\%) & 1.0 & \multicolumn{1}{l|}{$<$ 0.001} & 10 (2.8\%) & 1.0 & $<$ 0.001 \\
      \multicolumn{1}{l|}{SOC} & 41 (10.3\%) & $<$ 0.001 & \multicolumn{1}{l|}{1.0} & 45 (11.3\%) & $<$ 0.001 & 1.0 \\
      \multicolumn{1}{l|}{MSP} & 72 (18.0\%) & $<$ 0.001 & \multicolumn{1}{l|}{0.002} & 75 (18.8\%) & $<$ 0.001 & 0.003 \\
      \bottomrule
    \end{tabular}
    \label{tab:msp_vs_soc_vs_rand}
  \end{table*}

%% file: tables/algo_runtimes.tex
\begin{table}[t]
    \caption{Time to compute importance scores of text blocks of size $B=10$ tokens.  For SOC we sampled 100 contexts per block from a 10-block radius.  For MSP the expected 
    number of times a given block was masked was 100. Runtimes were averaged over 20 randomly sampled discharge summaries.}
    \resizebox{\columnwidth}{!}{%
    \begin{tabular}{lll}
    \toprule
      \textbf{Algorithm} & \textbf{Mean (Stdv.) Runtime} \\
      \midrule
      \midrule
      SOC  & 17.81 (6.05) hours \\ 
      MSP ($P=0.1$)  & 0.89 (0.05) hours \\
      MSP ($P=0.5$)  & 0.18 (0.01) hours \\
      \bottomrule
    \end{tabular}
    \label{tab:algo_runtimes}
    }
  \end{table}

%% file: tables/algo_runtimes_theoretical.tex
\begin{table}[t]
    \caption{Required model inferences to compute importance scores of text block pairs sampling 100 contexts per block with SOC and 
    setting $J=100$ for MSP.}
    \resizebox{\columnwidth}{!}{%
    \begin{tabular}{lll}
    \toprule
      \textbf{Algorithm} & \textbf{1000 Tokens} & \textbf{10,000 Tokens} \\
      \midrule
      \midrule
      SOC & 100,000,000 & 10,000,000,000 \\ 
      MSP ($P=0.1$) & 10,000 & 10,000 \\
      MSP ($P=0.5$) & 400 & 400 \\
      \bottomrule
    \end{tabular}
    \label{tab:algo_runtimes_theoretical}
    }
  \end{table}

%% file: tables/model_perf_compare.tex
\begin{table*}
\centering
\caption{
    Summary of performance across two datasets. 
    For comprehensive metrics, see Supplemental Tables~\ref{tab:640000_perf_transpose}, \ref{tab:8067_perf_transpose}, and \ref{tab:bert_mimic_results}.
    In parentheses we show the model sequence length for CAML and Big Bird and the aggregation method used with RoBERTa. 
    The pretraining column specifies the type of text the model (or embeddings for CAML) was pretrained with (general or clinical).
    Mean (standard deviation) micro- and macro-average precision values shown.
    LR: logistic regression; AP: average precision (i.e., AUPR).
}
\label{tab:model_perf_compare}
\begin{tabular}{p{2.0cm}|llll}
\toprule
\textbf{Dataset}                                    & \textbf{Method}            & \textbf{Pretraining} & \textbf{Micro-AP} & \textbf{Macro-AP} \\ \hline \hline
\multirow{3}{*}{\optumshortname} & TF-IDF + LR       & Clinical               & 0.6391 (0.0010)         & 0.2541 (0.0013)         \\
                                           & CAML (32,768)             & Clinical               & 0.8550 (0.0008)         & 0.5796 (0.0026)         \\
                                           & Big Bird (32,768)         & Clinical               & 0.9087 (0.0005)         & 0.6461 (0.0027)         \\ \hline \hline
\multirow{7}{*}{MIMIC}              & TF-IDF + LR     & Clinical              & 0.4574 (0.0058)        & 0.3791 (0.0052)        \\
                                           & RoBERTa (max)        & General                     & 0.6482 (0.0056)         & 0.5447 (0.0057)         \\
                                           & CAML (5,000)      & Clinical              & 0.6950 (0.0044)          & 0.6101 (0.0052)         \\
                                           & Big Bird (4,096)  & General              & 0.6663 (0.0059)          & 0.5648 (0.0058)         \\
                                           & Big Bird (4,096)  & Clinical               & 0.6998 (0.0053)         & 0.6138 (0.0053)         \\
                                           & Big Bird (32,768) & Clinical             & 0.6927 (0.0052)          & 0.6103 (0.0053)          \\ 
\bottomrule
\end{tabular}
\end{table*}

%% file: tables/descriptive_stats.tex
\begin{table*}[h]
    \centering 
    \caption{Descriptive statistics for training and evaluation datasets.}
    \begin{tabular}{llll}
    \toprule
       & \textbf{Optum} & \textbf{MIMIC} \\
      \midrule
      \midrule
      \# Documents & 6,526,116 & 11,371 \\ 
      Median (IQR) tokens per document & 4043 [1830-9142] & 1429.5 [1029-1929] \\
      Median (IQR) labels per document & 1 [0-3] & 5 [3-8] \\
      Total \# of labels & 85 & 50 \\
      \bottomrule
    \end{tabular}
    \label{tab:descriptive_stats} 
\end{table*}

%% file: tables/big_bird_params.tex
\begin{table*}[ht]
    \caption{Hyperparameters used for original Big Bird model and our Big Bird implementation. Differences are shown in bold.}
\begin{tabular}{p{6cm}p{3cm}p{5cm}}
\hline
Parameter                                    & \begin{tabular}[c]{@{}l@{}}Big Bird Paper \\  (MLM Pretraining\\ Section E)\end{tabular} & \begin{tabular}[c]{@{}l@{}}Big Bird for Medical Documents\\ (this work) \end{tabular}                  \\ \hline
Subword Token Vocab Size                     & 32,000                                                                                      & 32,000                                            \\
Max Position Embedding Size                  & 4,096                                                                                       & \textbf{32,768}                                            \\
Hidden Size                                  & 768                                                                                         & 768                                               \\
Intermediate Size                            & 3,072                                                                                       & 3,072                                             \\
Number of Hidden Layers                      & 12                                                                                          & 12                                                \\
Number of Attention Heads                    & 12                                                                                          & 12                                                \\
Token Block Size (Sliding, Random,   Global) & 64                                                                                          & 64                                                \\
Number of Random Blocks                      & 3                                                                                           & 3                                                 \\
Hidden Dropout Probability                   & 0.1                                                                                         & 0.1                                               \\
Attention Dropout Probability                & 0.1                                                                                         & 0.1                                               \\
Activation Layer                             & GELU                                                                                        & GELU                                              \\
Optimizer                                    & Adam                                                                                        & Adam                                              \\
Loss                                         & Cross-entropy                                                                               & Cross-entropy                                     \\
Learning Rate                                & 0.0001                                                                                      & 0.0001                                            \\
Batch Size                                   & 256                                                                                         & \textbf{32 (effective batch size) \& 4 accumulation steps} \\
Hardware                                     & 8 x 8 TPU                                                                                   & \textbf{4 x 8 GPU}                                         \\
Warmup Steps                                 & 10,000                                                                                      & 10,000                                            \\ \hline
\end{tabular}
\label{tab:big_bird_params}
\end{table*}

%% file: tables/12800_perf_transpose.tex
\begin{table*}
\centering
\caption{
Performance of text classifiers trained on 12,800 chart sample of the \optumname dataset averaged across 85 conditions.  
Performance was averaged over 100 bootstrap iterations on the test dataset.
Standard deviations are reported in parentheses.
}
\label{tab:12800_perf_transpose}
\begin{tabular}{p{3.1cm}p{2.7cm}p{2.7cm}p{2.7cm}p{2.7cm}}
\toprule
Text Representation  & TF-IDF & TF-IDF & Word2Vec (pretrained)  &                            Big Bird LM (pretrained)  \\
\midrule
Classifier  &  One-vs-Rest ElasticNet Logistic Regression  &  Feed Forward Neural Net  &                  CAML  &  Fine-Tuned Big Bird Model with Classification Head  \\
\midrule
 Micro-PR AUC   &                             0.1709 (0.0007)  &          0.4482 (0.0009)  &       0.7422 (0.0010)  &                                    0.8122 (0.0008)  \\
 Macro-PR AUC   &                             0.0355 (0.0003)  &          0.1167 (0.0006)  &       0.3615 (0.0013)  &                                    0.4044 (0.0013)  \\
 Micro-ROC AUC  &                             0.8825 (0.0003)  &          0.9155 (0.0003)  &       0.9687 (0.0002)  &                                   0.9750 (0.0002)  \\
 Macro-ROC AUC  &                             0.5505 (0.0008)  &          0.7195 (0.0019)  &       0.8524 (0.0015)  &                                   0.9101 (0.0012)  \\
 Micro-F1       &                             0.0012 (0.0001)  &          0.3610 (0.0012)  &       0.7230 (0.0007)  &                                   0.7870 (0.0006)  \\
 Macro-F1       &                            0.0001 (0.0000)   &         0.0656 (0.0004)   &      0.3726 (0.0011)   &                                  0.3274 (0.0007)   \\
\bottomrule
\end{tabular}
\end{table*}

%% file: tables/64000_perf_transpose.tex
\begin{table*}
\centering
\caption{
Performance of text classifiers trained on 64,000 chart sample of the \optumname dataset averaged across 85 conditions.  
Performance was averaged over 100 bootstrap iterations on the test dataset.
Standard deviations are reported in parentheses.
}
\label{tab:64000_perf_transpose}
\begin{tabular}{p{3.1cm}p{2.7cm}p{2.7cm}p{2.7cm}p{2.7cm}}
\toprule
Text Representation  & TF-IDF & TF-IDF & Word2Vec (pretrained)  &                            Big Bird LM (pretrained)  \\
\midrule
Classifier  &  One-vs-Rest Elasticnet Logistic Regression  &  Feed Forward Neural Net  &                  CAML  &  Fine-Tuned Big Bird Model with Classification Head  \\
\midrule
 Micro-PR AUC   &                             0.5363 (0.0009)  &          0.5787 (0.0011)  &       0.8139 (0.0008)  &                                   0.8689 (0.0006)  \\
 Macro-PR AUC   &                             0.1694 (0.0007)  &          0.1994 (0.0008)  &       0.4965 (0.0022)  &                                   0.5263 (0.0017)  \\
 Micro-ROC AUC  &                             0.9275 (0.0003)  &          0.9353 (0.0003)  &       0.9827 (0.0001)  &                                   0.9869 (0.0001)  \\
 Macro-ROC AUC  &                             0.7247 (0.0013)  &          0.8011 (0.0016)  &       0.9428 (0.0014)  &                                   0.9590 (0.0006)  \\
 Micro-F1       &                             0.3958 (0.0011)  &          0.5373 (0.0010)  &       0.7682 (0.0008)  &                                   0.8300 (0.0006)  \\
 Macro-F1       &                            0.0886 (0.0005)   &         0.1466 (0.0007)   &      0.4677 (0.0019)   &                                  0.4658 (0.0013)   \\
\bottomrule
\end{tabular}
\end{table*}

%% file: tables/128000_perf_transpose.tex
\begin{table*}
\centering
\caption{
Performance of text classifiers trained on 128,000 chart sample of the \optumname dataset averaged across 85 conditions.  
Performance was averaged over 100 bootstrap iterations on the test dataset.
Standard deviations are reported in parentheses.
}
\label{tab:128000_perf_transpose}
\begin{tabular}{p{3.1cm}p{2.7cm}p{2.7cm}p{2.7cm}p{2.7cm}}
\toprule
Text Representation  & TF-IDF & TF-IDF & Word2Vec (pretrained)  &                            Big Bird LM (pretrained)  \\
\midrule
Classifier  &  One-vs-Rest Elasticnet Logistic Regression  &  Feed Forward Neural Net  &                  CAML  &  Fine-Tuned Big Bird Model with Classification Head  \\
\midrule
 Micro-PR AUC   &                             0.5569 (0.0011)  &          0.5139 (0.0010)  &       0.8296 (0.0009)  &                                   0.8800 (0.0006)  \\
 Macro-PR AUC   &                             0.1836 (0.0008)  &          0.1404 (0.0006)  &       0.5402 (0.0026)  &                                   0.5627 (0.0018)  \\
 Micro-ROC AUC  &                             0.9346 (0.0003)  &          0.9275 (0.0003)  &       0.9869 (0.0001)  &                                   0.9868 (0.0001)  \\
 Macro-ROC AUC  &                             0.7741 (0.0016)  &          0.7830 (0.0016)  &       0.9538 (0.0011)  &                                   0.9612 (0.0009)  \\
 Micro-F1       &                             0.4259 (0.0010)  &          0.4824 (0.0008)  &       0.7909 (0.0006)  &                                   0.8362 (0.0006)  \\
 Macro-F1       &                            0.1007 (0.0005)   &         0.0909 (0.0003)   &      0.5553 (0.0024)   &                                  0.4827 (0.0013)   \\
\bottomrule
\end{tabular}
\end{table*}

%% file: tables/640000_perf_transpose.tex
\begin{table*}
    \centering
    \caption{
    Performance of text classifiers trained on 640,000 chart sample of the \optumname dataset averaged across 85 conditions.  
    Performance was averaged over 100 bootstrap iterations on the test dataset.
    Standard deviations are reported in parentheses.
    }
    \label{tab:640000_perf_transpose}
    \begin{tabular}{p{3.1cm}p{2.7cm}p{2.7cm}p{2.7cm}p{2.7cm}}
    \toprule
    Text Representation  & TF-IDF & TF-IDF & Word2Vec (pretrained)  &                            Big Bird LM (pretrained)  \\
    \midrule
    Classifier  &  One-vs-Rest ElasticNet Logistic Regression  &  Feed Forward Neural Net  &                  CAML  &  Fine-Tuned Big Bird Model with Classification Head  \\
    \midrule
     Micro-PR AUC   &                             0.6391 (0.0010)  &          0.5397 (0.0009)  &       0.8550 (0.0008)  &                                    0.9087 (0.0005)  \\
     Macro-PR AUC   &                             0.2541 (0.0013)  &          0.1609 (0.0008)  &       0.5796 (0.0026)  &                                    0.6461 (0.0027)  \\
     Micro-ROC AUC  &                             0.9479 (0.0002)  &          0.9349 (0.0003)  &       0.9887 (0.0001)  &                                   0.9927 (0.0001)  \\
     Macro-ROC AUC  &                             0.8179 (0.0019)  &          0.8097 (0.0017)  &       0.9622 (0.0011)  &                                   0.9789 (0.0006)  \\
     Micro-F1       &                             0.5855 (0.0009)  &          0.4853 (0.0011)  &       0.8025 (0.0006)  &                                   0.8521 (0.0006)  \\
     Macro-F1       &                            0.2183 (0.0010)   &         0.1034 (0.0005)   &      0.5726 (0.0026)   &                                  0.5725 (0.0020)   \\
    \bottomrule
    \end{tabular}
    \end{table*}

%% file: tables/8067_perf_transpose.tex
\begin{table*}
\centering
\caption{
Performance of text classifiers trained on 8,067 discharge summaries from the MIMIC dataset averaged across 50 conditions.  
Performance was averaged over 100 bootstrap iterations on the test dataset.
Standard deviations are reported in parentheses.
The CAML paper reports p-values but no measures of variation and does not include Micro-Average PR AUCs.
The clinically pretrained Big Bird models are pretrained on the \optumname dataset with maximum sequence lengths of 32,768 tokens 
and 4,096 tokens while the generically pretrained Big Bird model is the model from \citet{zaheer2021big} with a maximum sequence length of 4,096 tokens.
OvR: one-versus-rest; LR: logistic regression; FFNN: feed-forward neural network; A.D.S.: all discharge summaries; C.P.T.: clinical pretraining; G.P.T.: generic pretraining;
MSL: max sequence length; TR: text representation
}
\label{tab:8067_perf_transpose}
\begin{tabular}{p{2.0cm}p{1.5cm}p{1.5cm}p{1.5cm}p{1.5cm}p{1.5cm}p{1.5cm}p{1.5cm}p{1.5cm}}
\toprule
MSL  &          Full  &            Full  &     2,500 words  &        2,500 words  &      5,000 words  &    32,768 subwords  &     4,096 subwords  &     4,096 subwords  \\
\midrule
TR  &           TF-IDF  &           TF-IDF  &  Word2Vec (A.D.S.)  &  Word2Vec (C.P.T.)  &  Word2Vec (C.P.T.)  &  Big Bird (C.P.T.)  &  Big Bird (G.P.T.) &  Big Bird (C.P.T.)  \\
\midrule 
Classifier  &      OvR LR  &           FFNN  &    Original CAML  &          CAML  &             CAML  &         Big Bird  &       Big Bird       &         Big Bird  \\
\midrule
 Micro-PR AUC   &    0.4574 (0.0058)  &    0.4964 (0.0060)  &             N/A  &    0.6913 (0.0047)  &    0.6950 (0.0044)  &    0.6927 (0.0052)  &    0.6663 (0.0059)  &  0.6998 (0.0053) \\
 Macro-PR AUC   &    0.3791 (0.0052)  &    0.4043 (0.0051)  &              N/A  &    0.6031 (0.0052)  &    0.6101 (0.0052)  &    0.6103 (0.0053)  &    0.5648 (0.0058)  &  0.6138 (0.0053) \\
 Micro-ROC AUC  &    0.8068 (0.0026)  &    0.8264 (0.0027)  &             0.909  &    0.9205 (0.0017)  &    0.9246 (0.0016)  &    0.9174 (0.0019)  &    0.9042 (0.0024) & 0.9211 (0.0019) \\
 Macro-ROC AUC  &    0.7697 (0.0032)  &    0.7809 (0.0033)  &            0.875  &    0.8946 (0.0023)  &    0.8979 (0.0023)  &    0.8902 (0.0025)  &    0.8723 (0.0027)  & 0.8947 (0.0023) \\
 Micro-F1       &    0.2250 (0.0066)  &    0.3972 (0.0062)  &            0.614  &    0.6333 (0.0046)  &    0.6378 (0.0044)  &    0.6016 (0.0047)  &    0.6037 (0.0051)  & 0.6529 (0.0046) \\
 Macro-F1       &   0.1184 (0.0028)   &   0.2825 (0.0049)   &           0.532   &   0.5413 (0.0049)   &   0.5410 (0.0046)   &   0.4780 (0.0048)   &   0.4727 (0.0050)   & 0.5532 (0.0049) \\
\bottomrule
\end{tabular}
\end{table*}

%% file: tables/bert_mimic_results.tex
\begin{table*}
\centering
\caption{
Performance of Transformer-based text classifiers trained on 8,067 discharge summaries from the MIMIC dataset averaged across 50 conditions.  
Performance was averaged over 100 bootstrap iterations on the test dataset.
Standard deviations are reported in parentheses.
For the Bio-ClinicalBERT \citep{huang_clinicalbert_2020,alsentzer_publicly_2019} and the RoBERTa \citep{liu_roberta_2019} models, 
the model's max sequence length is shorter than the document length. 
Therefore, we evaluated several different methods for handling longer input sequences: truncation of the sequence to the model's limit (base) and
aggregating model outputs over windows of tokens using several different functions (mean, max, custom). 
The custom aggregation function is described in equation 4 of \citet{huang_clinicalbert_2020}.
Big Bird has a maximum sequence length of 4,096 tokens and we compare Big Bird with generic pretraining (G.P.T) using MLM as described in \citet{zaheer2021big}
to Big Bird with clinical pretraining (C.P.T.).
C.B.: Bio-ClinicalBERT \citep{huang_clinicalbert_2020,alsentzer_publicly_2019}; R.B.: RoBERTa \citep{liu_roberta_2019}; B.B.: Big Bird \citep{zaheer2021big};
}
\label{tab:bert_mimic_results}
\begin{tabular}{p{1.5cm}p{1.2cm}p{1.2cm}p{1.2cm}p{1.2cm}p{1.2cm}p{1.2cm}p{1.2cm}p{1.2cm}p{1.2cm}}
\toprule
Max Seq Length & 512 & Any & Any & Any & 512 & Any & Any & 4096 & 4096 \\
\midrule
{} &      C.B. (base) &      C.B. (mean) &       C.B. (max) &    C.B. (custom) &      R.B. (base) &      R.B. (mean) &       R.B. (max) &      B.B. (G.P.T) &   B.B.  (C.P.T)\\
\midrule
Micro AP    &  0.5532 (0.0062) &  0.6333 (0.0059) &  0.6765 (0.0052) &  0.6431 (0.0059) &  0.4904 (0.0064) &  0.6179 (0.0059) &  0.6482 (0.0056) &  0.6663 (0.0059) & 0.6998 (0.0053) \\
Macro AP    &  0.4533 (0.0058) &  0.5714 (0.0060) &  0.5834 (0.0053) &  0.5909 (0.0056) &  0.3917 (0.0053) &  0.5552 (0.0057) &  0.5447 (0.0057) &  0.5648 (0.0058) & 0.6138 (0.0053) \\
Micro AUROC &  0.8461 (0.0031) &  0.9096 (0.0019) &  0.9104 (0.0019) &  0.9111 (0.0019) &  0.8158 (0.0033) &  0.9012 (0.0020) &  0.8964 (0.0023) &  0.9042 (0.0024) & 0.9211 (0.0019) \\
Macro AUROC &  0.8098 (0.0036) &  0.8864 (0.0023) &  0.8787 (0.0024) &  0.8879 (0.0023) &  0.7680 (0.0038) &  0.8776 (0.0025) &  0.8585 (0.0028) &  0.8723 (0.0027) & 0.8947 (0.0023) \\
Micro F1    &  0.4931 (0.0055) &  0.4654 (0.0062) &  0.6282 (0.0047) &  0.4614 (0.0060) &  0.4012 (0.0059) &  0.4083 (0.0066) &  0.6133 (0.0047) &  0.6037 (0.0051) & 0.6529 (0.0046) \\
Macro F1    &  0.3629 (0.0050) &  0.3324 (0.0059) &  0.5154 (0.0049) &  0.3273 (0.0058) &  0.2457 (0.0044) &  0.2629 (0.0051) &  0.4836 (0.0045) &  0.4727 (0.0050) & 0.5532 (0.0049) \\
\bottomrule
\end{tabular}
\end{table*}

%% file: tables/optumcharts85_performance_compare.tex
\onecolumn
\begin{longtable}{p{7.5cm}lll}
\caption{
  Per-label area under the precision-recall (AUPR) curve values for the CAML model and the Big Bird model evaluated on the \optumname test dataset.
  Models were trained on 640,000 samples from the \optumname dataset.
  Prevalence is the fraction of samples where the label occurred.
  Results are sorted by prevalence in ascending order.  
  }
\label{tab:optumcharts85_performance_compare}\\
\toprule
                                         Condition & Prevalence & CAML AUPR & Big Bird AUPR \\
\midrule
\endfirsthead

\toprule
                                         Condition & Prevalence & CAML AUPR & Big Bird AUPR \\
\midrule
\endhead
\midrule
\multicolumn{4}{r}{{Continued on next page}} \\
\midrule
\endfoot

\bottomrule
\endlastfoot
Pressure Ulcer of Skin with Necrosis Through to... &      $<$0.001 &     0.036 &         0.230 \\
Amyotrophic Lateral Sclerosis and Other Motor N... &      $<$0.001 &     0.307 &         0.267 \\
        Substance Use with Psychotic Complications &      $<$0.001 &     0.355 &         0.147 \\
             Coma, Brain Compression/Anoxic Damage &      $<$0.001 &     0.125 &         0.100 \\
                                Muscular Dystrophy &      $<$0.001 &     0.596 &         0.662 \\
                                      Quadriplegia &      0.001 &     0.544 &         0.524 \\
         Respirator Dependence/Tracheostomy Status &      0.001 &     0.282 &         0.335 \\
Pressure Ulcer of Skin with Full Thickness Skin... &      0.001 &     0.180 &         0.256 \\
                                 Major Head Injury &      0.001 &     0.166 &         0.083 \\
     Aspiration and Specified Bacterial Pneumonias &      0.001 &     0.254 &         0.103 \\
                           Intracranial Hemorrhage &      0.001 &     0.167 &         0.154 \\
Pressure Ulcer of Skin with Partial Thickness S... &      0.001 &     0.224 &         0.258 \\
                          Opportunistic Infections &      0.001 &     0.226 &         0.264 \\
                             Personality Disorders &      0.001 &     0.353 &         0.400 \\
                 Diabetes with Acute Complications &      0.001 &     0.279 &         0.377 \\
                                        Paraplegia &      0.001 &     0.471 &         0.577 \\
                          Hip Fracture/Dislocation &      0.001 &     0.308 &         0.328 \\
Substance Use Disorder, Mild, Except Alcohol an... &      0.001 &     0.249 &         0.316 \\
             Monoplegia, Other Paralytic Syndromes &      0.001 &     0.147 &         0.179 \\
                                    Cerebral Palsy &      0.001 &     0.788 &         0.885 \\
                         Unspecified Renal Failure &      0.002 &     0.197 &         0.310 \\
Atherosclerosis of the Extremities with Ulcerat... &      0.002 &     0.469 &         0.475 \\
Complications of Specified Implanted Device or ... &      0.002 &     0.157 &         0.269 \\
    Vertebral Fractures without Spinal Cord Injury &      0.002 &     0.191 &         0.377 \\
                              Chronic Pancreatitis &      0.002 &     0.682 &         0.784 \\
      Major Organ Transplant or Replacement Status &      0.002 &     0.651 &         0.686 \\
                Intestinal Obstruction/Perforation &      0.002 &     0.277 &         0.388 \\
                    Severe Hematological Disorders &      0.002 &     0.564 &         0.702 \\
                    Spinal Cord Disorders/Injuries &      0.003 &     0.344 &         0.419 \\
     Pneumococcal Pneumonia, Empyema, Lung Abscess &      0.003 &     0.367 &         0.417 \\
                           End-Stage Liver Disease &      0.003 &     0.712 &         0.805 \\
             Bone/Joint/Muscle Infections/Necrosis &      0.003 &     0.425 &         0.563 \\
Pressure Pre-Ulcer Skin Changes or Unspecified ... &      0.003 &     0.549 &         0.671 \\
Unstable Angina and Other Acute Ischemic Heart ... &      0.003 &     0.289 &         0.361 \\
                                          HIV/AIDS &      0.003 &     0.738 &         0.859 \\
Septicemia, Sepsis, Systemic Inflammatory Respo... &      0.004 &     0.339 &         0.516 \\
                       Dementia With Complications &      0.004 &     0.713 &         0.736 \\
                      Protein-Calorie Malnutrition &      0.004 &     0.555 &         0.692 \\
                       Acute Myocardial Infarction &      0.004 &     0.211 &         0.343 \\
Proliferative Diabetic Retinopathy and Vitreous... &      0.005 &     0.733 &         0.838 \\
    Artificial Openings for Feeding or Elimination &      0.005 &     0.513 &         0.621 \\
                             Disorders of Immunity &      0.006 &     0.527 &         0.658 \\
                                     Schizophrenia &      0.006 &     0.823 &         0.903 \\
                                Multiple Sclerosis &      0.006 &     0.817 &         0.856 \\
              Metastatic Cancer and Acute Leukemia &      0.006 &     0.613 &         0.663 \\
                                 Chronic Hepatitis &      0.007 &     0.777 &         0.841 \\
                    Ischemic or Unspecified Stroke &      0.007 &     0.141 &         0.190 \\
                                   Dialysis Status &      0.007 &     0.774 &         0.836 \\
                    Exudative Macular Degeneration &      0.007 &     0.894 &         0.920 \\
                                Cirrhosis of Liver &      0.008 &     0.826 &         0.913 \\
               Vascular Disease with Complications &      0.008 &     0.303 &         0.402 \\
Amputation Status, Lower Limb/Amputation Compli... &      0.008 &     0.747 &         0.831 \\
                            Hemiplegia/Hemiparesis &      0.008 &     0.685 &         0.773 \\
                   Chronic Kidney Disease, Stage 5 &      0.009 &     0.801 &         0.856 \\
                        Inflammatory Bowel Disease &      0.009 &     0.826 &         0.895 \\
                     Lung and Other Severe Cancers &      0.010 &     0.704 &         0.796 \\
                               Acute Renal Failure &      0.011 &     0.479 &         0.603 \\
              Cardio-Respiratory Failure and Shock &      0.011 &     0.701 &         0.830 \\
            Colorectal, Bladder, and Other Cancers &      0.011 &     0.490 &         0.624 \\
            Chronic Ulcer of Skin, Except Pressure &      0.012 &     0.701 &         0.794 \\
          Chronic Kidney Disease, Severe (Stage 4) &      0.012 &     0.831 &         0.904 \\
 Fibrosis of Lung and Other Chronic Lung Disorders &      0.013 &     0.773 &         0.838 \\
             Parkinson's and Huntington's Diseases &      0.013 &     0.826 &         0.916 \\
                        Lymphoma and Other Cancers &      0.015 &     0.760 &         0.840 \\
Substance Use Disorder, Moderate/Severe, or Sub... &      0.016 &     0.723 &         0.802 \\
                                         Nephritis &      0.018 &     0.689 &         0.772 \\
                                   Angina Pectoris &      0.021 &     0.724 &         0.819 \\
Other Significant Endocrine and Metabolic Disor... &      0.022 &     0.742 &         0.839 \\
Coagulation Defects and Other Specified Hematol... &      0.024 &     0.696 &         0.809 \\
                 Seizure Disorders and Convulsions &      0.025 &     0.822 &         0.896 \\
                     Dementia Without Complication &      0.031 &     0.837 &         0.903 \\
    Breast, Prostate, and Other Cancers and Tumors &      0.042 &     0.673 &         0.764 \\
 Major Depressive, Bipolar, and Paranoid Disorders &      0.053 &     0.880 &         0.922 \\
                Reactive and Unspecified Psychosis &      0.053 &     0.877 &         0.921 \\
Rheumatoid Arthritis and Inflammatory Connectiv... &      0.058 &     0.863 &         0.911 \\
        Chronic Kidney Disease, Moderate (Stage 3) &      0.066 &     0.928 &         0.960 \\
                                    Morbid Obesity &      0.079 &     0.899 &         0.941 \\
                          Congestive Heart Failure &      0.099 &     0.899 &         0.936 \\
Myasthenia Gravis/Myoneural Disorders and Guill... &      0.108 &     0.890 &         0.936 \\
                                  Vascular Disease &      0.110 &     0.858 &         0.906 \\
Chronic Kidney Disease, Mild or Unspecified (St... &      0.112 &     0.913 &         0.958 \\
                       Specified Heart Arrhythmias &      0.113 &     0.934 &         0.964 \\
             Chronic Obstructive Pulmonary Disease &      0.121 &     0.933 &         0.960 \\
               Diabetes with Chronic Complications &      0.182 &     0.888 &         0.953 \\
                     Diabetes without Complication &      0.263 &     0.911 &         0.950 \\
\end{longtable}
\twocolumn

%% file: tables/mimic50_performance_compare.tex
\onecolumn
\begin{longtable}{p{7.5cm}lll}
\caption{
  Per-label area under the precision-recall (AUPR) curve values for the CAML model and the Big Bird model evaluated on the MIMIC test dataset.
  Models were trained using the MIMIC training data.
  Prevalence is the fraction of samples where the label occurred.
  Results are sorted by prevalence in ascending order.  
  }
\label{tab:mimic50_performance_compare}\\
\toprule
                                         Condition & Prevalence & CAML AUPR & Big Bird AUPR \\
\midrule
\endfirsthead

\toprule
                                         Condition & Prevalence & CAML AUPR & Big Bird AUPR \\
\midrule
\endhead
\midrule
\multicolumn{4}{r}{{Continued on next page}} \\
\midrule
\endfoot

\bottomrule
\endlastfoot
                       Transfusion of packed cells &      0.029 &     0.218 &         0.142 \\
Combined right and left heart cardiac catheteri... &      0.035 &     0.343 &         0.397 \\
                                      Hemodialysis &      0.042 &     0.800 &         0.910 \\
                    Diagnostic ultrasound of heart &      0.049 &     0.358 &         0.301 \\
                Other endoscopy of small intestine &      0.050 &     0.567 &         0.633 \\
Parenteral infusion of concentrated nutritional... &      0.053 &     0.734 &         0.747 \\
Acute myocardial infarction, subendocardial inf... &      0.053 &     0.497 &         0.666 \\
                      Unspecified pleural effusion &      0.055 &     0.317 &         0.358 \\
                Left heart cardiac catheterization &      0.058 &     0.588 &         0.600 \\
  Pneumonitis due to inhalation of food or vomitus &      0.059 &     0.504 &         0.601 \\
                            Mitral valve disorders &      0.060 &     0.598 &         0.551 \\
          Postsurgical aortocoronary bypass status &      0.062 &     0.734 &         0.797 \\
                         Old myocardial infarction &      0.070 &     0.314 &         0.417 \\
            Closed [endoscopic] biopsy of bronchus &      0.073 &     0.640 &         0.510 \\
    Single internal mammary-coronary artery bypass &      0.078 &     0.964 &         0.956 \\
                     Thrombocytopenia, unspecified &      0.080 &     0.429 &         0.280 \\
                          Arterial catheterization &      0.086 &     0.336 &         0.301 \\
                            Unspecified septicemia &      0.086 &     0.481 &         0.468 \\
                 Hyposmolality and/or hyponatremia &      0.090 &     0.475 &         0.471 \\
                         Pure hypercholesterolemia &      0.090 &     0.530 &         0.556 \\
        Coronary arteriography using two catheters &      0.091 &     0.855 &         0.845 \\
Chronic airway obstruction, not elsewhere class... &      0.093 &     0.625 &         0.664 \\
                                     Severe sepsis &      0.095 &     0.651 &         0.679 \\
               Chronic kidney disease, unspecified &      0.099 &     0.502 &         0.458 \\
                   Pneumonia, organism unspecified &      0.101 &     0.541 &         0.628 \\
Encounter for long-term (current) use of antico... &      0.105 &     0.702 &         0.667 \\
                              Tobacco use disorder &      0.105 &     0.378 &         0.199 \\
                            History of tobacco use &      0.108 &     0.263 &         0.163 \\
Continuous mechanical ventilation for 96 consec... &      0.108 &     0.711 &         0.730 \\
                                          Acidosis &      0.113 &     0.451 &         0.410 \\
     Depressive disorder, not elsewhere classified &      0.113 &     0.533 &         0.550 \\
                      Acute posthemorrhagic anemia &      0.117 &     0.604 &         0.649 \\
                        Unspecified hypothyroidism &      0.121 &     0.799 &         0.855 \\
Hypertensive renal disease, unspecified, withou... &      0.124 &     0.731 &         0.701 \\
Extracorporeal circulation auxiliary to open he... &      0.131 &     0.973 &         0.979 \\
Enteral infusion of concentrated nutritional su... &      0.132 &     0.660 &         0.657 \\
                    Insertion of endotracheal tube &      0.135 &     0.542 &         0.587 \\
                               Anemia, unspecified &      0.143 &     0.183 &         0.217 \\
       Urinary tract infection, site not specified &      0.145 &     0.693 &         0.713 \\
                         Acute respiratory failure &      0.147 &     0.690 &         0.672 \\
Continuous mechanical ventilation for less than... &      0.149 &     0.607 &         0.642 \\
                                 Esophageal reflux &      0.154 &     0.722 &         0.762 \\
type II diabetes mellitus [non-insulin dependen... &      0.197 &     0.748 &         0.744 \\
                  Acute renal failure, unspecified &      0.209 &     0.645 &         0.656 \\
  Venous catheterization, not elsewhere classified &      0.233 &     0.549 &         0.537 \\
             Congestive heart failure, unspecified &      0.244 &     0.863 &         0.883 \\
Coronary atherosclerosis of native coronary artery &      0.252 &     0.864 &         0.910 \\
                               Atrial fibrillation &      0.272 &     0.916 &         0.925 \\
              Other and unspecified hyperlipidemia &      0.317 &     0.728 &         0.776 \\
                Unspecified essential hypertension &      0.450 &     0.798 &         0.840 \\
\end{longtable}
\twocolumn

%% file: tables/masking_experiment.tex
\begin{table*}[t]
  \centering
  \caption{Number and proportion of informative text blocks (IBs) for making a randomly sampled diagnosis from MIMIC discharge summaries.
  Text blocks of size $K=10$ tokens were randomly masked over $N$ iterations with masking probability $P$ before running inference with Big Bird.
  We set the number of iterations $N$ for each experiment to $1000/P$ such that the expected number of times a given text block
  is masked when computing importance is equal to 1000.  For example, for $P=0.1$, we set $N=10,000$.
  The p-value comes from a two-tailed, two-sample T-test without assuming equal variances, comparing the proportion of informative
  blocks between those chosen through the masking procedure for a given $P$ and blocks chosen at random.}
  \begin{tabular}{llll}
  \toprule
    \textbf{Masking Probability} & \textbf{Count of IBs} & \textbf{Proportion of IBs}  & \textbf{p-value} \\
    \midrule
    \midrule
    0.1  & 35 & 0.299 & $<$ 0.001 \\
    0.3  & 27 & 0.231 & $<$ 0.001 \\
    0.5  & 26 & 0.222 & $<$ 0.001 \\
    0.9  & 10 & 0.085 & 0.302 \\
    Random & 6 & 0.051 & 1.000 \\
    \bottomrule
  \end{tabular}
  \label{tab:masking_experiment}
\end{table*}

%% file: tables/algo_runtimes_len.tex
\begin{table*}[t]
    \centering
    \caption{Below are mean runtimes over 20 experiments for each text block importance algorithm on documents of various fixed lengths.
    Standard deviation is reported in parentheses. We compare our masked sampling procedure (MSP) at two masking probabilities $P$ to the Sampling and Occlusion
    (SOC) algorithm \citep{jin_towards_2020}. Note the rapid increase in SOC runtimes, even at these modest document lengths,
    making SOC intractable for very long documents.}
    \begin{tabular}{lllll}
    \toprule
      \textbf{Algorithm} & \textbf{50 Token Doc} & \textbf{100 Token Doc} & \textbf{500 Token Doc} & \textbf{1,000 Token Doc} \\
      \midrule
      \midrule
      SOC  & 0.31 (0.05) mins & 0.47 (0.04) mins & 2.17 (0.03) mins & 65.49 (0.59) mins \\
      MSP ($P=0.1$)  & 0.27 (0.03) mins & 0.26 (0.03) mins & 0.33 (0.02) mins & 6.31 (0.08) mins \\
      MSP ($P=0.5$)  & 0.11 (0.02) mins & 0.12 (0.02) mins & 0.12 (0.02) mins & 1.41 (0.04) mins \\
      \bottomrule
    \end{tabular}
    \label{tab:algo_runtimes_len}
  \end{table*}